\documentclass[10pt,twocolumn,letterpaper]{article}

\usepackage{times}
\usepackage{epsfig}
\usepackage{graphicx}
\usepackage{float}
\usepackage{wrapfig}
\usepackage{amsmath,amssymb,amsthm}
\usepackage{algorithm,algorithmicx,algpseudocode}

\usepackage{bm,xspace}
\usepackage{comment}
\usepackage{verbatim}
\usepackage{multirow}
\usepackage{balance}
\usepackage{url}
\usepackage{booktabs}
\usepackage{etoolbox,siunitx}
\usepackage{calc}
\usepackage{pifont,hologo}
\usepackage[usenames, dvipsnames]{color}
\usepackage{nicefrac}
\usepackage{cvpr}
\usepackage{color}
\usepackage{xcolor}
\usepackage{tcolorbox}
\usepackage{appendix}
\usepackage{gensymb}  

\usepackage{paralist}

\theoremstyle{definition}

\DeclareMathSymbol{@}{\mathord}{letters}{"3B}

\usepackage{macros/mysymbols}

\newcommand{\secref}[1]{Sec.~\ref{#1}\xspace}

\newcommand{\bs}{\boldsymbol}

     {\begin{list}{}%
             {\setlength{\leftmargin}{#1}}%
             \item[]%
     }
{\end{list}}

\newcommand{\myquote}[1]{\emph{`#1'}}
\newcommand{\mypred}[1]{\texttt{\footnotesize #1}}

\definecolor{citecolor}{HTML}{2980b9}
\definecolor{linkcolor}{HTML}{c0392b}
\usepackage[pagebackref=true,breaklinks=true,colorlinks,bookmarks=false,citecolor=citecolor,linkcolor=linkcolor]{hyperref}
\usepackage{cleveref}

\cvprfinalcopy 

\makeatletter
\def\blfootnote{\xdef\@thefnmark{}\@footnotetext}
\makeatother

\makeatletter
\def\and{%
  \end{tabular}%
  \hskip 2em \@plus.17fil\relax
  \begin{tabular}[t]{c}}
\makeatother

\begin{document}

\title{Rearrangement: A Challenge for Embodied AI}

\author{Dhruv Batra\\
{\small Georgia Tech}\\{\small Facebook AI Research}\\{}
\and
Angel X. Chang\\
{\small Simon Fraser University}
\and
Sonia Chernova\\
{\small Georgia Tech}\\{\small Facebook AI Research}\\{}
\and
Andrew J. Davison\\
{\small Imperial College London}
\and
Jia Deng\\
{\small Princeton University}
\and
Vladlen Koltun\\
{\small Intel Labs}
\and
Sergey Levine\\
{\small UC Berkeley}\\{\small Google}
\and
Jitendra Malik\\
{\small UC Berkeley}\\{\small Facebook AI Research}\\{}
\and
Igor Mordatch\\
{\small Google}
\and
Roozbeh Mottaghi\\
{\small Allen Institute for AI}\\{\small University of Washington}\\{\vspace{-3mm}}
\and
Manolis Savva\\
{\small Simon Fraser University}
\and
Hao Su\\
{\small UC San Diego}
}

\maketitle

\begin{abstract}

We describe a framework for research and evaluation in Embodied AI. Our proposal is based on a canonical task: Rearrangement. A standard task can focus the development of new techniques and serve as a source of trained models that can be transferred to other settings. In the rearrangement task, the goal is to bring a given physical environment into a specified state. The goal state can be specified by object poses, by images, by a description in language, or by letting the agent experience the environment in the goal state. We characterize rearrangement scenarios along different axes and describe metrics for benchmarking rearrangement performance. To facilitate research and exploration, we present experimental testbeds of rearrangement scenarios in four different simulation environments. We anticipate that other datasets will be released and new simulation platforms will be built to support training of rearrangement agents and their deployment on physical systems.

\end{abstract}

\blfootnote{The authors are listed in alphabetical order.}

\section{Introduction}
\label{sec:introduction}

Embodied AI is the study and development of intelligent systems with a physical or virtual embodiment.
Over the past few years, significant advances have been made in developing intelligent agents that can navigate in previously unseen environments. These advances have been accelerated by dedicated software platforms~\cite{Kolve2017,Savva2017,Xia2018,Savva2019} and clear experimental protocols~\cite{Anderson2018,Batra2020}. Navigation research is thriving in part due to healthy infrastructure and experimental methodology.

\begin{figure*}
    \includegraphics[width=\textwidth]{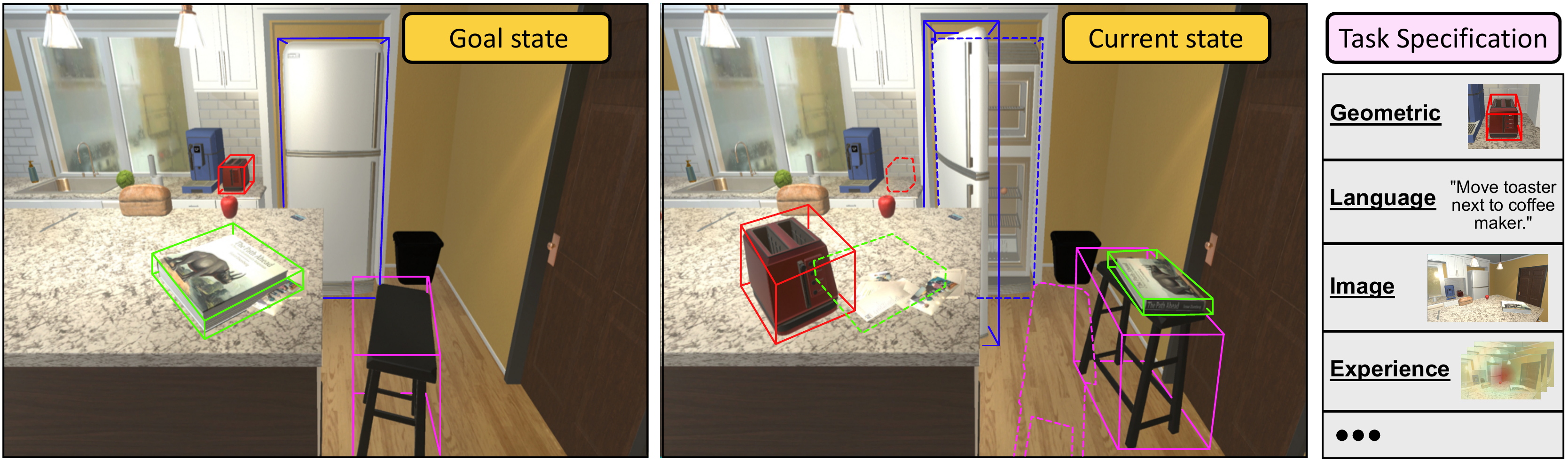}
    \caption{\textbf{Object rearrangement example.} The goal and the current state of the scene are shown in the left and right images, respectively. The agent is required to move objects (e.g., chair) or change their state (e.g., close the fridge) to recover the goal configuration. The rightmost panel shows different ways of specifying the rearrangement task. }
    \label{fig:example}
\end{figure*}

An exciting frontier for Embodied AI research concerns interaction and contact between the agent and the environment: tasks that call on the agent to actively engage with and modify the environment in order to accomplish its goals. A number of software platforms support such interaction scenarios~\cite{James2020,Kolve2017,Savva2019,Xiang2020,Gan2020}. These software platforms simulate realistic onboard perception and in some cases the physical dynamics of the agent, environment, and their interaction.

One missing ingredient is a clear task definition that can span different software platforms and catalyze coordinated accumulation of knowledge and ability across research groups. Clear task definitions and evaluation metrics are essential in the common task framework, which is substantially responsible for progress in computer vision and natural language processing~\cite{Donoho2017}.

In computer vision, standard tasks such as image classification and object detection have facilitated the development of foundational techniques and architectures that have enriched the whole field~\cite{Everingham2015,Russakovsky2015}. Language modeling and machine translation have served a similar role in natural language processing~\cite{Sutskever2011,Sutskever2014,Bahdanau2015,Vaswani2017}. In both fields, these standard tasks focus the development and validation of new representations and algorithms, and serve as a source of trained models that can be transferred to other tasks, as with convolutional backbones pretrained for image classification~\cite{He2016}, object detectors~\cite{Ren2015,He2020}, and transformers pretrained on natural language~\cite{Devlin2019,Radford2019,Yang2019,Brown2020}.

\begin{figure*}
    \includegraphics[width=\textwidth]{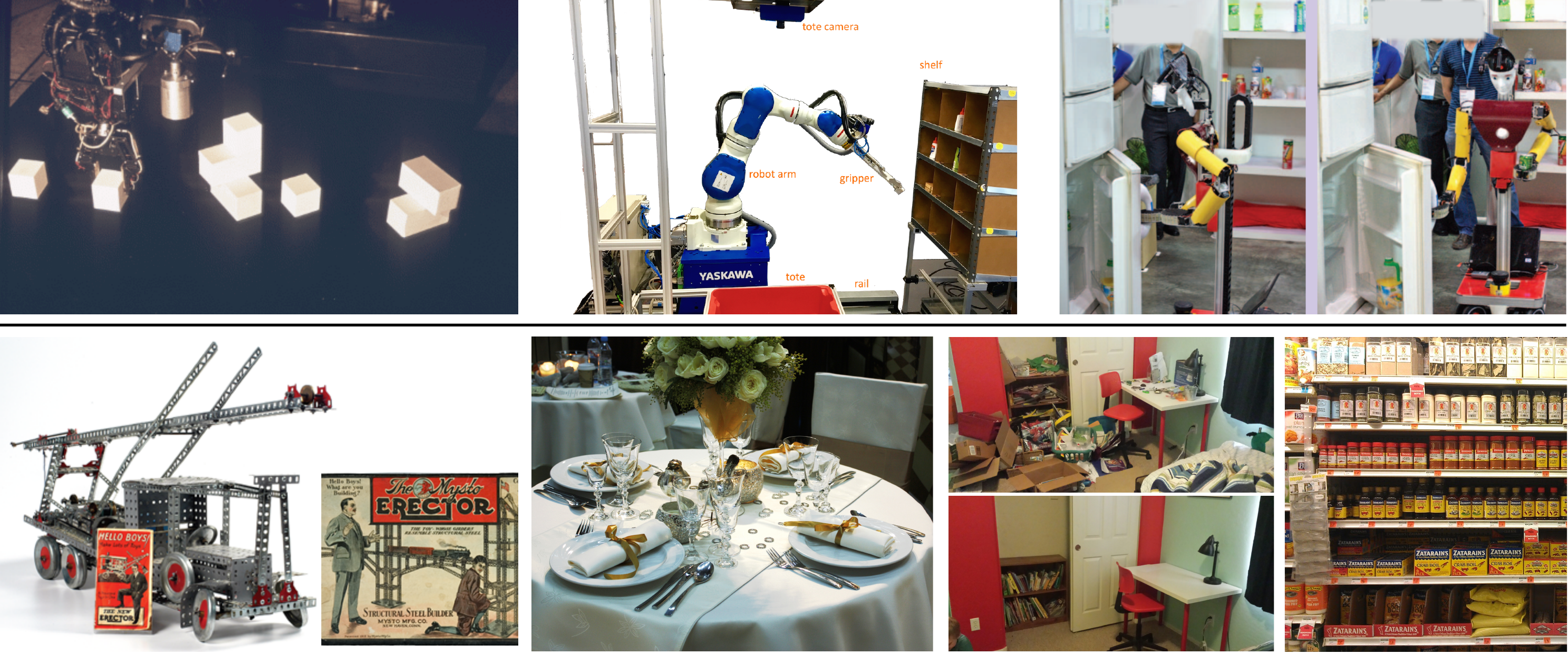}
    \caption{\textbf{Examples of rearrangement tasks.} Top row: examples of experimental settings in vision, robotics, and artificial intelligence that can be viewed as rearrangement tasks. From left to right: the MIT copy demo is a seminal demonstration of robotic rearrangement; the Amazon Robotics Challenge involves moving items from a shelf into a tote (image from Hernandez et al.~\cite{hernandez2016team}); RoboCup@Home is a long-running robotics competition involving household tasks (image from Stuckler et al.~\cite{stuckler2012robocup}). Bottom row: people perform many rearrangement tasks that remain challenging for current robotic systems. From left to right: construction toys are a popular activity for children (image from \url{toyhalloffame.org}); setting a dining table, cleaning up a bedroom, stocking grocery shelves are all rearrangements performed by people.}
    \label{fig:examples}
\end{figure*}

In this report, we develop a task definition that can likewise align and accelerate research in Embodied AI. The task is rearrangement: Given a physical environment, bring it into a specified goal state. Figure~\ref{fig:example} provides an example. We propose rearrangement as a canonical task for Embodied AI because it naturally unifies instances that are of clear practical interest: setting the table, cleaning the bedroom, loading the dishwasher, picking and placing orders in a fulfillment center, rearranging the furniture, and many more. Rearrangement scenarios can be defined with stationary manipulators that operate locally or with mobile systems that traverse complex scenes such as houses and apartments. Many experimental settings that have been explored in robotics can be viewed as instances of rearrangement, as well as many compelling settings that are beyond the reach of present-day systems (Figure~\ref{fig:examples}).

We focus on rearrangement of rigid and articulated piecewise-rigid objects. This includes a broad range of interesting and challenging scenarios, as illustrated in Figure~\ref{fig:examples}.
In order to focus research on settings that we consider challenging but tractable in the near future,
we deliberately exclude within-object transformations, such as melting (spot welding, soldering), destruction (cutting, drilling, sanding), and non-rigid dynamics (liquids, films, cloths, or ropes).
Thus boiling water, chopping onions, stomping grapes, pouring coffee, making a smoothie, folding towels, ironing clothes,
or making a bed are considered beyond the scope of the presented framework.
We expect these
scenarios to be incorporated as technology matures in the future.

Rearrangement calls upon a variety of abilities in an embodied intelligent system. Successful rearrangement can require recognizing the state of objects in the environment (is the cabinet open or closed?), inferring the differences between the current and the goal state, manipulating objects in cluttered environments (e.g., an object cannot be moved if it is blocked by another object), estimating forces required to move objects and predicting the effect of those forces, planning a sequence of actions, navigating through complex environments while maintaining a persistent internal representation, and learning the pre-conditions and post-conditions of each action. Rearrangement is thus a comprehensive framework for research and evaluation in Embodied AI that subsumes component skills such as navigation and manipulation and integrates them into a broader roadmap towards embodied intelligence.

The goal state of rearrangement can be specified in different forms: geometric configuration, images of the desired state, natural language, a formal predicate-based specification, or an embodied specification that lets the system examine the environment in the goal state before being instantiated in the initial configuration. We define an evaluation protocol that can accommodate all of these specifications by evaluating predicates that score objects and object sets in terms of their compliance to the goal. A rearrangement episode receives a score between 0 and 1, enabling clear ranking of competing approaches.

We provide a number of guidelines in order to maximize the practical utility of this research program and assist subsequent deployment on physical systems. First, we emphasize the importance of acting based on perceptual input: sensing the environment with realistic sensors, rather than planning in an idealized map, configuration space, or logical abstraction. Embodied AI puts noisy and incomplete onboard perception in the control loop: this makes the development of robust systems challenging but is necessary for successful adoption in the real world. Our second key recommendation is to work either in the physical world or in physical dynamics simulation. Objects should be moved by forces, most commonly applied via contact. Lastly, we advocate prioritizing strong generalization: evaluating rearrangement agents with objects and environments that were never encountered during training, with no access to privileged information such as 3D models, ground-truth object poses, or perfect localization. Agents should be able to perform rearrangement tasks in previously unseen environments, the content and layout of which are accessibly solely through realistic onboard sensing.

To assist near-term research and development, we release suites of rearrangement scenarios in four simulation environments: THOR~\cite{Kolve2017}, RLBench~\cite{James2020}, SAPIEN~\cite{Xiang2020}, and Habitat~\cite{Savva2019}. We anticipate that other scenario datasets will be released and new simulation platforms will be built to support this line of work. We hope that the task specification, evaluation protocol, and broader discussion in this report will support healthy development of Embodied AI and the creation of intelligent systems that perceive, act, and accomplish increasingly long-term goals in complex physical environments.

\section{Background}
\label{sec:background}

Rearrangement can be approached in a modular way, for example through a decomposition into submodules such as perception and planning. Here perception converts sensory input to a representation of the world, from which planning produces actions. Planning has been the subject of a long line of work in Artificial Intelligence~\cite{Russell1995ArtificialI}. Going back to the early days of AI, Newell and Simon~\cite{Simon1962ComputerSO} proposed GPS, the General Problem Solver, which could be related to even more classical work, the means-ends analysis of Aristotle. More concretely, the Shakey robot project~\cite{Kuipers2017} at SRI in the late 1960s led to the STRIPS~\cite{fikes1971strips} formulation of a plan as applying operators in sequence with each operator having its precondition, add and delete lists expressed in a logical formalism. The representation language used by STRIPS for specifying planning problems has been replaced by PDDL (Problem Domain Description Language), introduced by Ghallab et al.~\cite{pddl}, and there is a flourishing community with an annual conference and challenge competitions. A number of books and surveys cover this body of work~\cite{geffner13,Ghallab2016,KarpasMagazzeni2020}. This style of planning is sometimes called task planning to distinguish it from motion planning, which has been the subject of extensive study in robotics. The focus in motion planning is much more geometric~-- planning paths for moving objects while avoiding obstacles in configuration spaces. LaValle~\cite{LaValle2006} is a good entry into this literature. The combination of task planning and motion planning has also been explored, for example by Kaelbling and Lozano-P\'{e}rez~\cite{Kaelbling2011HierarchicalTA,garrett2020integrated}.

In the task and motion planning literature, there has been a  rich body of work on `rearrangement planning'~\cite{ben1998practical,cosgun2011push,danielczuk2019mechanical,dogar2014object,King2016,King2017,krontiris2014rearranging,scholz2010combining,shome2020synchronized,stilman2007manipulation}, the problem of searching for a sequence of actions that transform an initial configuration of objects into some goal configuration. A variety of problem formulations exist, with different kinds of permitted transformations, such as pushing versus grasping. Such rearrangement planners usually do not directly address the problem of perception, in the sense that they do not deal with raw sensory input such as pixels; instead, they typically assume that the shape and pose of objects are already provided, possibly partially and with uncertainty.

Our point of departure from the classic task and motion planning literature is our emphasis on standardized end-to-end evaluation. End-to-end evaluation measures the capability of the full system from raw sensory input to actuation. Such evaluation is agnostic to the choice of approach and internal decomposition, putting more focus on real-world perception and action, as with physical sensors and actuators on robots and in physical simulation.

We believe that standardized end-to-end evaluation is synergistic with research on task and motion planning. Many of the existing task and motion planning methods are not easily comparable to each other due to different input/output assumptions. Standardized end-to-end evaluation can shed light on the role of planning techniques in the context of a full system that must handle real perception and actuation. It also allows comparing approaches that differ in design philosophy, such as modular systems based on classic planning techniques and monolithic neural networks that directly map sensory input to actuation commands.

Our work is of course not the first to present standardized evaluation protocols or benchmarks for Embodied AI problems. A number of efforts in robotics, computer vision, and machine learning have proposed various standardized evaluation methods, including protocols for evaluating robotic grasping~\cite{matheus2010benchmarking,ulbrich2011opengrasp,mahler2018guest}, navigation~\cite{Anderson2018,Batra2020},  manipulation tasks focusing on standardized real-world object sets~\cite{calli2015ycb}, and competition-style setups, such as RoboCup~\cite{kitano1997robocup,stone2005reinforcement} and the International Planning Competition~\cite{mcdermott20001998}, as well as simulated benchmarks focusing on specific algorithmic approaches such as reinforcement learning~\cite{bellemare2013arcade,brockman2016openai,fan2018surreal,yu2020meta,fu2020d4rl}.
Our proposal addresses a need that we believe is not met by these prior works.

Our aim is to develop a framework and protocol for evaluating Embodied AI systems that is general enough to cover a wide range of different capabilities (as compared, for example, to more narrow evaluations focusing on specific tasks, such as grasping) and can compare a large variety of algorithmic approaches on the same footing.
We believe that such a framework for evaluation of Embodied AI systems will empower researchers to pursue general and ambitious goals in terms of capabilities, without constraining to a specific and narrow range of approaches.

\section{Rearrangement}
\label{sec:rearrangement}

In this section, we provide a concrete and general but application-grounded definition for rearrangement. This is followed by
recommendations on agent embodiment and sensor specification. Evaluation criteria are discussed in \secref{sec:evaluation}.

\paragraph{Notation.}
For a set $\calS$, we will use $2^\calS$ to denote the power set (the set of all possible subsets) of $\calS$.
We specify Rearrangement using the language and notation of Partially Observable Markov Decision Processes (POMDP) because
this is a familiar and convenient mathematical abstraction. We do not however take any stance on approaches used to solve the problem.
Let $s \in \calS$ denote a state and state space,
$o \in \calO$ denote an observation and observation space,
$a \in \calA$ denote an action and action space,
$\calT(s_t,a_t,s_{t+1}) = Pr(s_{t+1} \, | \, s_t,a_t)$
denote the transition probability,
$g \in \calG$ denote a goal and goal space, and
$\pi(a_t; g) = Pr(a_t \mid o_0,\ldots,o_{t-1},a_0,\ldots,a_{t-1},g)$ denote the agent's goal-conditioned policy.

\paragraph{Articulated Rigid-body POMDPs.}
We will restrict our attention to worlds comprising a collection of rigid bodies and
kinematic trees thereof. Specifically, robots and objects are modeled as a tree
of rigid parts (legs, wheels, \etc) connected via joints that determine degrees of freedom
(see Section 3.3.2 in \cite{LaValle2006} for a refresher).
Let $SO(3)$ denote the special orthogonal group, the space of 3D rotations,
and $SE(3) = \RF^3 \times SO(3)$ denote the special Euclidean group, the space of rigid-body poses (3D locations and rotations).
In Rearrangement, the world state space is
factorized -- \ie, can be written as the Cartesian product of the rigid-body pose spaces corresponding to each of the $n$ parts:
\begin{subequations}
\begin{align}
 & \calS = \calS_1 \times \calS_2 \ldots \calS_n  \\
 \text{where} \qquad & \calS_i = SE(3) = \RF^3 \times SO(3) \qquad \forall i.
\end{align}
\end{subequations}
Notice that the expression above does not
account for the constraints imposed by the joints in the kinematic chains/trees and thus not all configurations in this
state space may be achievable.
Finally, we note that the above state-space specification ostensibly appears to exclude a number of
important variables -- any form of dynamics (velocities, acceleration, \etc), physical properties (mass, friction coefficients, \etc), or any
notion of time at all. However, as we describe next, the only role of this state space is to specify the problem
(in terms of the initial state, goal state(s), and the goal specifications). We do not take any stand on the intermediate states
that a solution to this problem may pass through.

\paragraph{Initial and Goal State(s).}
In Rearrangement, we will be concerned about two special states --
\begin{asparaenum}
\item an initial or starting state $s_0$ where the agent and environment find themselves at the beginning of the task,
and
\item a desired goal state $s^*$ that the agent attempts to rearrange the environment into.
Importantly, $s^*$ may not be unique, but is rather an element from a set of \emph{acceptable}
goal states, \ie $s^* \in S^* \subseteq \calS$. The exact specification of $S^*$ will depend
on the task at hand, but one convenient characterization may be via a finite set of predicates $P_i(\cdot)$. Thus, $S^*$ is the set of states $s^*$ for which
all $P_i(s^*)$ hold true ($\forall i$).
\end{asparaenum}

\paragraph{Goal Specification.}
Let $\phi: \calS \times 2^\calS \mapsto \calG$ denote a goal-specification function. Specifically,
given a starting state $s_0$
and a set of acceptable goal states $S^*$,
this function generates the goal specification
$g = \phi(s_0,S^*)$ for the agent.
Since noisy and incomplete onboard perception is a first-class citizen in Embodied AI,
the agent will typically not have access
to the state space or
any goal state $s^*$.
Instead, the agent must operate solely from observations $o$ and goal specification $g$.
Note that this emphasis on partial visibility is only for the agent;
the `experiment designer' will typically have access to $S^*$ and use it
for evaluation (and potentially training of the agent).
This is a reasonable assumption for experiments conducted in simulation but may be infeasible for real hardware experiments.

\paragraph{Rearrangement Task Definition.}
With this notation in place, we can formally define Rearrangement.

\begin{tcolorbox}
Given a goal specification $g = \phi(s_0, S^*)$, an agent must transform an environment
from an initial starting state $s_0$ to a goal state $s^* \in S^*$, operating purely based on sensory observations $o \in \calO$.
\end{tcolorbox}

This abstract definition covers many aforementioned examples as special cases, instantiated by picking appropriate
choices of state, observation, and action space, and a goal specification. For instance, the state space in setting the table, cleaning the bedroom,
loading the dishwasher, picking and placing orders in a fulfillment center, and rearranging the furniture can all be defined (to a first degree of approximation)
as the product space of rigid-body pose spaces corresponding to each object.

\begin{figure*}[t]
    \centering
    \includegraphics[width=0.49\textwidth]{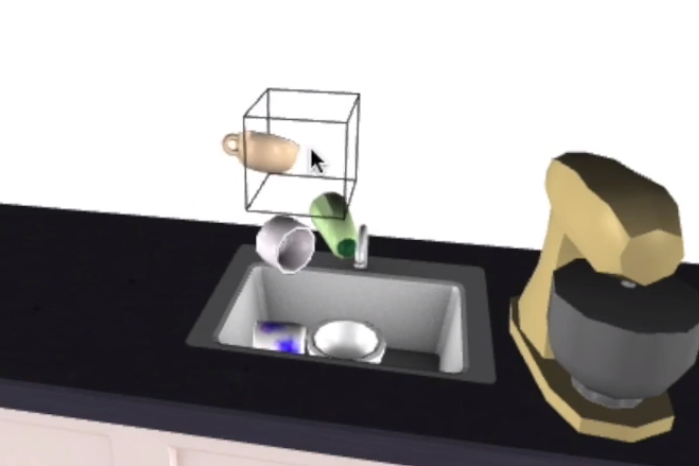}
    \includegraphics[width=0.49\textwidth]{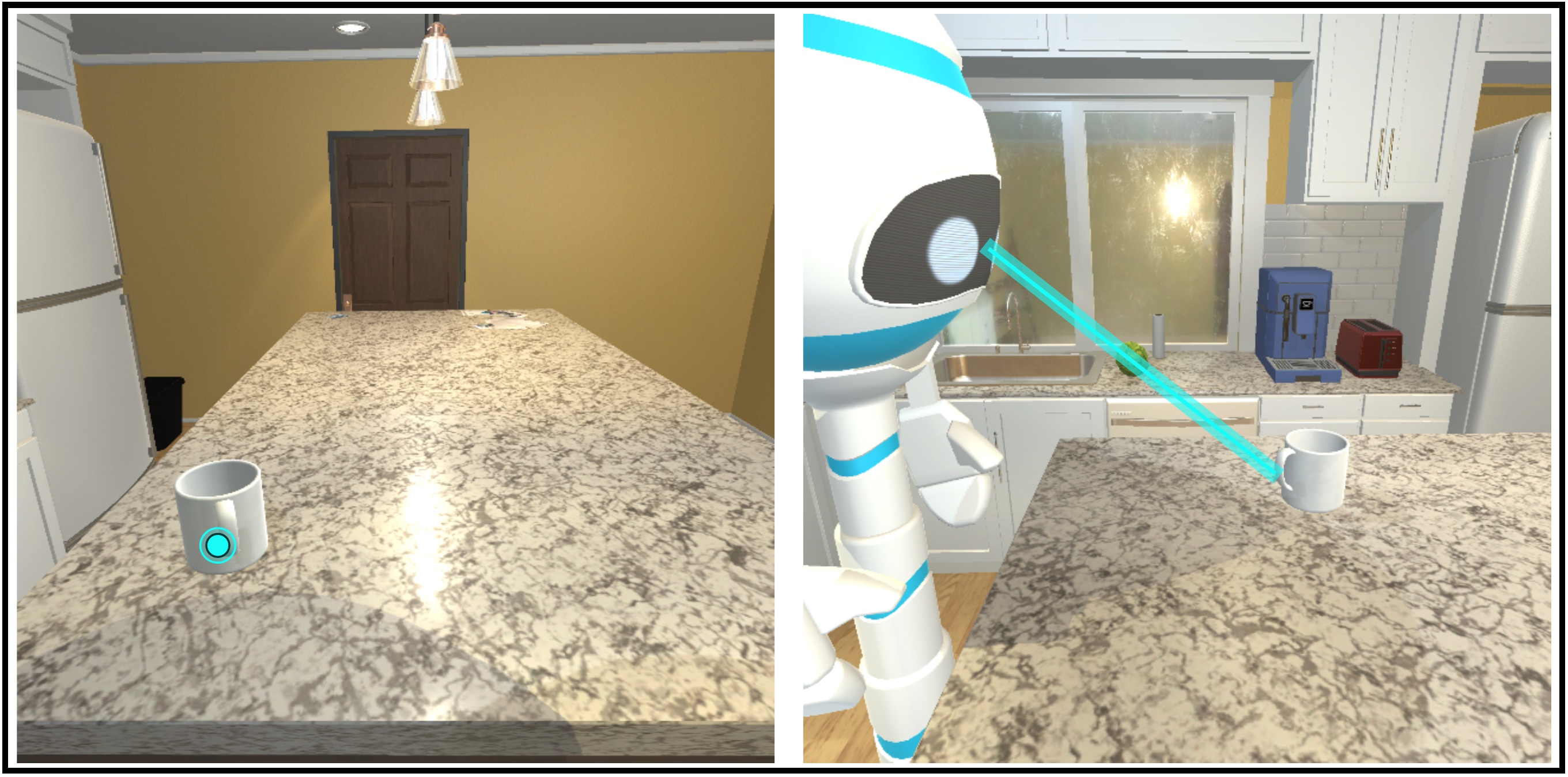}
    \caption{\textbf{Magic pointer abstraction} in Habitat~\cite{Savva2019} (left), using a mouse pointer, and AI2-THOR~\cite{Kolve2017} (right) using a raycast from the camera origin to the point of interaction (cyan markings are just for visualization purposes and not visible to the agent).}
    \label{fig:magic_pointer}
\end{figure*}

We now describe a number of goal specification mechanisms.
\begin{itemize}
\item \textbf{GeometricGoal.}
Consider a single object to be moved/rearranged.
In this setting, $g$ provides a geometric specification of the
transformation that this object undergoes from $s_0$ to
$s^*$.
This could be at various levels of detail --
(\eg the coordinates of the center of mass of the object in the goal state relative to the start state),
via 3D bounding box transformations
(\eg coordinates of an axis-aligned or oriented bounding box around
the object in the goal state in the coordinate system of the object in
the start state),
or via a full rigid-body pose transformations for articulated objects.
In multi-object rearrangement, $g$ can be a tuple of geometric configurations of each object to be moved. Note that $g$ is likely to be much lower-dimensional than $s$, \ie not all objects, agents, or places in the environment may be specified in the goal $g$.

\item \textbf{ImageGoal.}
In this setting, $g$ provides a visual rendering or representation of $s^*$, \eg
a 3rd person (say overhead, orthographic, perspective or isometric) image of the environment in the goal state.
While this is a convenient and informative way to specify the goal state -- simply by taking a picture -- it is important to note that a goal image may be underspecified. For instance, if certain objects are not visible in the image goal, there are multiple possible valid goal placements where they may be out of view. A goal image may also limit our ability to specify that we want certain objects to be placed or contained inside other objects.
One natural generalization of ImageGoal is VideoGoal, where the agent receives a sequence of images depicting the goal state. The camera pose corresponding to these images could be strategically chosen by the experiment
designer to disambiguate the underspecification associated with ImageGoal.

\item \textbf{LanguageGoal.}
In this setting, $g$ provides a linguistic description of the environment in the goal state (\eg \myquote{move the chair to the right of the sofa},
\myquote{move the blue block to the right edge of the table}). Note that language is typically underspecified and there may be several
goal states that fulfill a given language goal. This will require care in evaluation.

\item \textbf{ExperienceGoal.}
In this setting, we side-step the problem of designing a goal specification function $\phi(\cdot)$ by immersing the agent in the environment in
the goal condition $s^*$ (under a time/interaction budget) and letting the agent build whatever representation it deems appropriate
to bring the environment back to this state. Thus, the goal is essentially to \myquote{make it like it was before}. This captures the scenario
of asking a home robot to clean up the kitchen after a particularly messy session, where the robot has already experienced the kitchen in the clean state.
This idea may be generalized by also immersing the agent in `non-goal'
environments, so that it can distinguish between important and unimportant attributes of the environment.

\item \textbf{PredicateGoal.}
In this setting, $g$ is a set of predicates that should be satisfied in the goal state (\eg \mypred{on(plate1, table1)}).
A PredicateGoal can be precisely specified and evaluated by defining the predicates (\eg \mypred{supported\_by, inside, is\_on}), the symbols on which they act (\eg \mypred{plate, pizza, table, microwave}), and their grounding to objects and relations.
Symbols should map to objects or sets of objects, and each predicate should be evaluated by a function taking as input a state.
A predicate may be implemented in a number of ways -- geometric thresholding of positions of objects through a program, running a neural network classifier on the current state, or programatically checking the logical state of an object (\eg is the microwave on or off).
PredicateGoal forms a common substrate on top of which other goal types can be interpreted (\eg by conversion of a LanguageGoal to a PredicateGoal).
PredicateGoal is not necessarily a natural interface for humans, but it can be a useful interface for systems.
\end{itemize}

\subsection{Embodiment: Actuators and Sensors}

Manipulating and rearranging objects requires a simulated physical embodiment, leaving a number of open parameters, such as the action representation, the degree to which the manipulation interaction itself is simulated or abstracted away, and the particular capabilities afforded to the agent. The choice of embodiment in a given environment will inevitably influence the types of methods that are effective. We therefore do not prescribe a single specific embodiment, but instead recommend that benchmark and algorithm designers provide a clear and reproducible description, while we provide a brief discussion of several reasonable choices.
Generally, the action and perception representation is known to have a large impact on the performance of some types of algorithms, such as reinforcement learning, and being overly prescriptive with the embodiment may result in excessive focus on the particular challenges of a specific embodiment rather than the broader challenges inherent in Embodied AI. The particular choice of embodiment in a given environment will fall on a spectrum between the most abstract and the most realistic. We provide a few examples of points on this spectrum, starting with the most abstract and ending with the most concrete. Although we recommend using the most realistic actuator and sensor embodiments whenever possible, the more abstract representations may nonetheless present a useful simplification of the problem to enable more targeted algorithmic progress in specific areas.

\paragraph{Actuation: magic pointer abstraction.} One way to strike a compromise between abstraction and rich physical interaction is to borrow a commonly used metaphor in video games: instead of controlling a realistic simulated body, the agent navigates the environment and continuously controls its viewing direction (i.e., pitch and yaw) and, optionally, a virtual ``mouse pointer'' that can move on the screen (see Figure~\ref{fig:magic_pointer}). The agent can trigger a ``pick'' action that results in a ray cast a short distance in front of the agent, picking the closest object that intersects this cast ray. In the view direction abstraction (Figure~\ref{fig:magic_pointer}, right), the ray passes through the specified point of interaction, while in the case of a virtual mouse pointer (Figure~\ref{fig:magic_pointer}, left) it passes through the location of the mouse pointer on the screen. This object is now ``held'' by the agent, at which point it may be freely rotated and, with another action, placed back into the environment.
The release of the object may be simulated as a fall from a certain height
in front of the agent,
or being placed in the same pose (relative to the agent) as the object was
during ``pick-up''.
Optionally, this embodiment could provide an option for the agent to ``stow'' the currently held object into a virtual ``backpack'' (of fixed capacity).
In contrast to the ``discrete object grasping'' abstraction, this more complex embodiment allows more precise relocation of objects, for example in settings where we might want to reposition a variety of small items on a table. It also provides for the ability to carry out rudimentary physical interactions, since the object held in front of the agent can still interact physically with other objects in the environment. On the other hand, it also requires more precision: instead of simply standing in front of an object, the agent must choose a point on the image to interact with. This might be useful for example for inserting one object into another, rearranging deformable objects, or other physical interactions. However, this abstraction still omits most of the nuances of physical object interaction and controlling a simulated body. This abstraction is used in the AI2-THOR environment discussed in Section~\ref{sec:characterization}.

\paragraph{Actuation: kinematic articulated arm with abstracted grasping.} The next step along the spectrum from abstraction to realism is a simulated kinematic arm. In this setting, the agent controls a realistic model of a robotic arm, either via Cartesian end-effector control or joint space control (Cartesian 6-DoF end-effector should be sufficient in practice for most applications), but grasping of objects is abstracted away, such that any object that is located close enough to the end effector is automatically (virtually) ``grasped'' when the agent issues a ``pick'' command. This abstraction strikes a compromise between requiring control of an actual robotic body and avoiding some of the intricate dynamic complexities of object interaction, such as inertia and contact forces. This makes it appropriate for simplified manipulation scenarios and methods focusing on perception, while still providing some of the challenges and geometric constraints of a physical body.

\begin{figure}
    \includegraphics[width=\columnwidth]{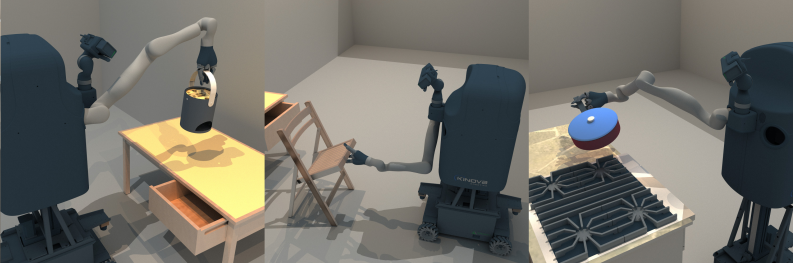}
    \caption{\textbf{Full physical simulation.} A simulated robot interacting with the environment in the SAPIEN~\cite{Xiang2020} framework. Full physical simulation provides the highest fidelity in terms of modeling object interaction, but also a more challenging setting necessitating continuous feedback control.}
    \label{fig:full_physical}
\end{figure}

\paragraph{Actuation: full physical simulation.} A full physical simulation of a robotic manipulator can provide the highest degree of fidelity (see Figure~\ref{fig:full_physical}), but also the greatest challenge in terms of requiring any algorithm to perform both effective closed-loop control and perception. Unlike the kinematic, abstracted grasping setting in the previous paragraph, a full physical simulation requires controlling a robotic arm to actually grasp objects physically, using accurately simulated contact forces. In this setting, the agent's actions directly map to actuation commands for simulated robotic joints. A number of choices must be made in instantiating such an embodiment: (1) end-effector Cartesian or joint-space arm control; (2) position control or force/torque control; (3) the type of gripper control afforded to the agent. A reasonable balance of fidelity and simplicity is to use (1) Cartesian-space control, (2) position control, (3) binary (open/close) gripper commands. However, other variants may also be used, and it is reasonable to consider this choice as part of the algorithm designer's prerogative, since all combinations are realistic for use on real-world robotic systems. A full physical simulation provides challenges that are most representative of those encountered in the real world, and is the only way to perform more nuanced physical object interaction. By simulating the low-level control stack (\eg PID or impedance control), appropriate actuation noise would also be simulated. This modality also poses additional perception challenges, since the robot must now determine not only which object to pick up, but also how to physically use its end-effector to pick up or manipulate the object successfully. Of course, this added complexity also places a heavier burden on algorithm designers. The SAPIEN and RLBench environments discussed in Section~\ref{sec:characterization} use this embodiment, as do most robotics simulation packages in use today.

\paragraph{Sensors: ground truth positions abstraction.} In some cases, researchers may choose to focus on planning and control without considering perception, in which case the coarsest abstraction of the perception problem is to utilize ground-truth positions and orientations of objects directly as input. This representation greatly simplifies the task and allows researchers to study the control problem in isolation. While we urge a combined and integrated effort to tackle perception and control jointly, we also recognize that some researchers may end up addressing these topics separately. In this case, reasonable representations could include positions and orientations of nearby objects, annotated with their identity and parameters of their geometric shape. Such positions should be provided with realistic noise, to ensure that any control or planning strategy is robust to perceptual uncertainty.

\paragraph{Sensors: intermediate representations.} An intermediate abstraction of onboard sensing is to assume mid-level visual processing, as would be typical in a modern perception stack, and directly utilize segmented images and/or depth maps~\cite{zhou2019does}. Many simulators can natively produce segmented images and depth maps, making this representation easy to obtain. This level of abstraction has a number of appealing properties: it allows researchers to avoid the need for training large pixel-level deep networks for processing raw pixels, abstracts away variation in lighting and appearance, and at the same time provides a realistic interface to current computer vision tools, since pixel-level segmentation and depth estimation are heavily studied topics. In this case, suitable noise should be added to the intermediate representations, including noise in the depth readings for depth maps and noise in the segmentation, including unlabeled pixels and pixels with erroneous labels.

\paragraph{Sensors: full simulated perception.} A full simulation of the robot's onboard sensors provides the highest fidelity of evaluation for sensing. In this case, simulated sensors might include RGB cameras, depth from RGBD sensors or LiDAR, as well as less common sensors, such as microphones and touch sensors. The sensors should be simulated with realistic noise and uncertainty and imperfect calibration (\eg camera calibration with small added noise). This representation provides the most realistic simulation of real-world deployment and we recommend this as the default mode of operation.

\subsection{Task Characterization}
\label{sec:characterization}

\begin{figure*}
\includegraphics[width=\textwidth]{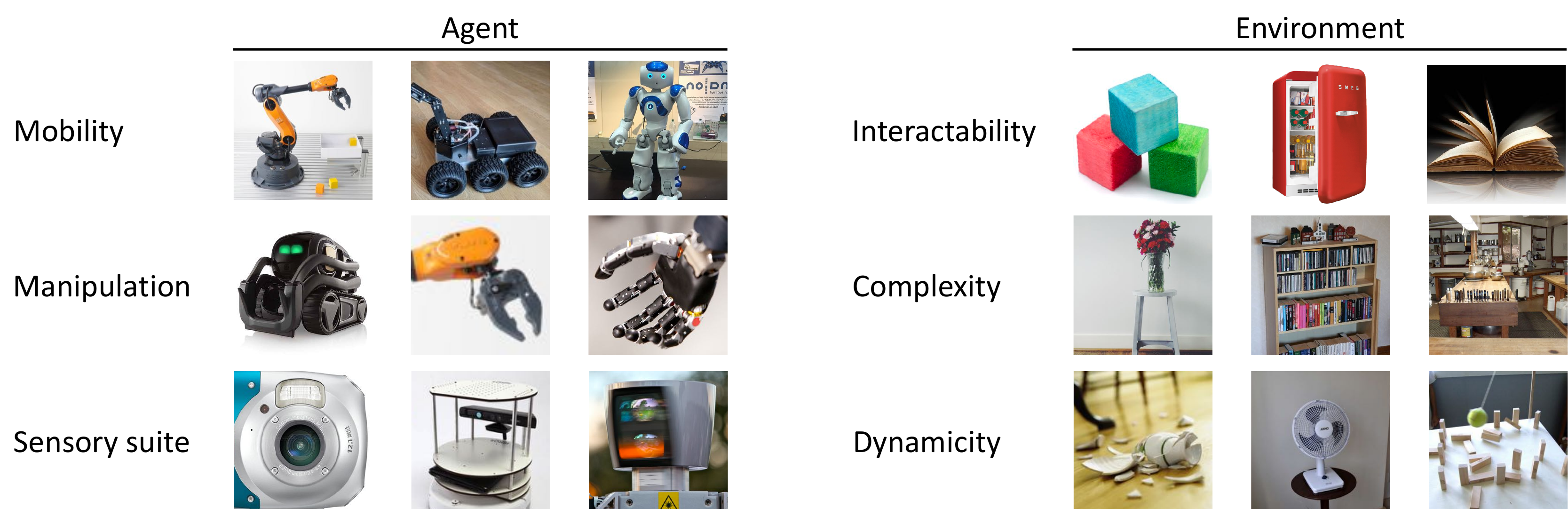}
\caption{\textbf{Dimensions of complexity characterizing the rearrangement task.} Several parameters of the agent and the environment determine the complexity of the rearrangement task. On the agent side: mobility (\eg fixed base, wheeled, bipedal), manipulator (\eg force applicator, parallel jaw gripper, humanoid hand), and sensory suite (\eg color camera, depth sensor, LiDAR). On the environment side: interactability (\eg rearrangement of rigid bodies, containers with rigid bodies, and articulated rigid bodies), complexity (\eg two objects, a full bookcase, a cluttered kitchen), and dynamicity (\eg object fracture, dynamic objects, and dynamic objects changing the state of other objects).}
\label{fig:characterization}
\end{figure*}

The complexity of a rearrangement task is characterized by several dimensions that can serve to define a taxonomy of rearrangement tasks.
We group these axes of complexity into agent-centric and environment-centric dimensions (see \Cref{fig:characterization}).

\paragraph{Agent mobility.}
Examples of agents that can perform rearrangement are fixed-base manipulator arms and mobile robots with manipulators.
The mobility characteristics of the agent restrict the rearrangement scenarios that can be performed (\eg a fixed robotic manipulator cannot move objects long distances).
The mobility of the agent also determines the available action space for navigation.
The action space can range from a set of discrete actions (`turn-left', `turn-right', `go-forward'), to parameterized actions, to fully continuous control for each motor. Although the locomotion problem itself can be made quite complex, with the addition of fully simulated legged locomotion and other intricacies, this dimension of the problem can also be reasonably abstracted away, for example by assuming that the robot's embodiment consists of a holonomic base. We will assume that mobility is accomplished by means of such a holonomic base for the purpose of this article.

\paragraph{Agent manipulation.}
Examples of manipulator types include: the ``magic pointer'' abstraction, ``sticky mittens''~\cite{needham2002pick}, suction-based grippers~\cite{brown2010universal}, needle/stick force applicators, parallel jaw grippers, and 5-finger humanoid grippers (see relevant survey in \cite{tai2016state}).
Earlier in this document, we elaborated on several concrete examples of agent manipulation (discrete grasping, magic pointer, and kinematic articulated arm).
We do not take a specific stance on manipulator types, though some manipulators are easier to simulate with current simulation platforms.
The spectrum of manipulation capabilities also forms a natural abstraction for research on rearrangement at higher levels (i.e., planning) vs.\ lower levels (i.e., control).

\paragraph{Agent sensory suite.}
We encourage broad investigation into sensors of different types and modalities.
A variety of sensor types are common: color vision cameras, depth sensors, contact/collision sensors, tactile sensors~\cite{martinez2016tactile}, LiDAR, microphones, and elastomeric sensors such as GelSight~\cite{johnson2009retrographic,yuan2017gelsight} are prominent examples.
Specific rearrangement scenarios may benefit from long-range visual sensing (\eg navigation-heavy rearrangement) or shorter-range haptic sensing (\eg single object grasping).
When using full simulated perception, we recommend that sensors should simulate noisy and incomplete onboard perception.
Odometry may be provided by the simulator, but with suitable noise and drift, necessitating realistic closed-loop corrections.

\paragraph{Environment interactability.}
The level of abstraction for the objects undergoing manipulation is another dimension of complexity.
In the simplest case, manipulated objects are rigid bodies with no additional state that the agent can manipulate.
Depending on the dexterity of the manipulator, the object may be rotated while it is held.
The agent may have a stowing capacity in which case the object can be stowed away from the manipulator.
A common abstraction is a virtual ``backpack'' with infinite or limited capacity stated in number of items, volume, or weight.
More complex scenarios may involve object articulation states (\eg books that may open and close).

\paragraph{Environment complexity.}
The difficulty of the rearrangement task depends on what is to be rearranged, the target configuration, and the structure of the environment.
Important parameters include the number of objects to be rearranged, the number of distractor objects, the degree of occlusion or containment in source and target configurations, whether the space the agent is moving in is relatively open or highly cluttered, and whether ordering is important (\eg stacking objects in order).
We focus on rearrangement of sets of piecewise rigid bodies in scenarios that may include containment and ordering constraints.
Note that this includes many common articulated objects such as cabinets.

\paragraph{Environment dynamicity.}
The degree to which the environment is dynamic forms another dimension of complexity.
Important parameters include whether the environment can change without the agent taking an action (\eg oscillating pedestal fan), whether there are unrecoverable states (\eg plates can break if dropped), and whether objects in the environment are subject to perturbations unrelated to the agent's actions (\eg wind or vibrations).

\subsection{Task Generalization Spectrum}

In addition to the above dimensions of task complexity, there exists a spectrum of task generalization settings.
Different points on this spectrum may involve generalization to novel objects, novel environments, novel source and target arrangements of the objects, as well as potentially novel agent actuation and sensing configurations.
On one extreme of the spectrum (`weak generalization') a system is evaluated with known objects in known environments, with only the arrangements of the objects being novel (i.e., there is a closed set of objects and environments shared between training and evaluation).
On the other side of the spectrum (`strong generalization'), the agent is tasked with rearranging new objects in new environments, never encountered during training.
We recommend strong generalization as the default mode of operation.

Regardless of where on the generalization spectrum a particular rearrangement problem falls, it is important to explicitly state the degree of overlap between prior experience or built-in knowledge (\eg object CAD models, grasp demonstrations for objects), and the test-time environment. It is also possible to allow limited exploration of an unseen test-time environment before the task commences.
An interesting question for future work lies in the quantification of these different forms of prior experience and their impact on task performance under different generalization settings.

\subsection{Comparisons and Evaluation}

The discussion in this section lays out a large number of potential decisions that can be made in evaluating an embodied system. From the choice of action abstraction, to the presence of dynamic and interactive components in the environment, these design choices create a spectrum of different levels of realism, and therefore a spectrum of difficulties. Naturally, any \emph{comparative} evaluation that aims to compare different methods must take these differing difficulty levels into account: a method that uses the \emph{magic pointer} abstraction is not directly comparable to one that uses a full physics simulation, and fair comparisons can only be performed at similar levels of abstraction and in similarly challenging environments. However, by reporting on these design dimensions and making clear the conditions under which each algorithm is evaluated, we can move closer to a setting where researchers can begin to put the performance of different methods in context, with evaluations that are at least conducted on tasks that vary along common axes of variation (action abstraction, sensor suite, interactability, complexity, and dynamicity).

\section{Evaluation}
\label{sec:evaluation}

A rearrangement task requires an agent (or agents) to accomplish a many-to-many transformation of a scene, and evaluation of success will necessarily therefore require the design of composite metrics which capture the quality of an overall performance. We believe that simplicity and the desire for rank-ordering make it valuable to focus on a single primary metric, and we propose that {\em task completion}, measured as a percentage, has high generality and flexibility for that purpose.
We define completion
as the percentage of binary target state tests concerning objects of interest which are successfully passed by the agent,
while not doing any harm to other parts of the scene.
We discuss the details required by this metric in this section.

Most tasks will have a natural range of difficulty among the objects which make them up, and so completion percentage will appropriately represent progression in agent performance.
As tasks increase in complexity, requiring more objects to be moved, this metric will become increasingly continuous.

Besides the primary completion metric, we strongly believe that
secondary metrics should also be reported, and in fact that these
will often be crucial in determining whether an agent has performance which
is on the path to useful Embodied AI applications.
While there are many possibilities,
we believe that the most important additional metrics should capture aspects of the agent's efficiency.
We make several concrete suggestions below.

\subsection{Primary Metric: Task Completion}

A rearrangement task is defined in terms of goal locations for movable objects in a simulated 3D environment.
As explained in Section~\ref{sec:rearrangement},
within our current scope we assume that all objects can be considered as rigid, or to consist of articulated rigid parts. Therefore the world state space $\calS = \calS_1 \times \calS_2 \ldots \calS_n $ is specified by the set of body pose spaces
$\calS_i = SE(3)$ for the $n$ objects or parts present in a scene. A particular current scene state is $s \in \calS$, and the goal state for a task is denoted $s^*$.
When an agent has finished working on a task, we must therefore determine our completion metric by comparing
the final state $s$ of all objects with goal state $s^*$.

The most straightforward approach is to fully decompose $s^*$
in terms of a target pose $s_i^* \in SE(3)$ for each individual object in absolute scene coordinates. The placement accuracy of each object would be evaluated via a norm $N(s_i, s_i^*)$
on the difference between each object's final and target pose.
The details of this norm can be designed to match the emphasis of a task:
\begin{itemize}
    \item  The norm could ignore rotation and be simply the translation between centers of mass.
\item The norm could combine translation distance  with a measure of rotation difference, for instance the angle between the two poses expressed in axis-angle form.
\item An elegant metric to unify translation and rotation which is familiar from 2D object detection is the 3D intersection-over-union (IoU) between the overlapping volumes or convex hulls.
\end{itemize}

Having decided on an error metric for individual objects, these values for multiple objects must be combined to obtain an overall score. It is tempting to average the individual object distances to obtain a mean accuracy measure. We believe that a better approach is to define a threshold on the distance for each object, make a binary test for each one, and then to report an overall completion percentage which is the fraction of objects passing the threshold tests. This completion metric can take account of the fact that tasks will be heterogeneous, and there may be very different sensible tolerances on the locations of different objects. For instance, placing a book anywhere on a table may be acceptable, whereas a flower must be placed within a narrow vase.

We believe that completion also has desirable robustness properties which make sense in terms of task progression, where one or two misplaced objects out of many will not have a ruinous impact on an overall score, even if those objects are very far from their target locations.

A crucial final issue to address with the completeness metric concerns its scope, because in most environments there could be many objects present which are not related to the task at hand. It seems clear that completion should only concern tests on objects of direct relevance, rather than all objects. However,
should we report an agent as successful if it achieves the correct state of the target objects but at the cost of making a mess of the rest of the scene, or even breaking other objects?
We argue no, since our goal is the development of methods with value in real-world robotics, and a useful robot must understand the scope of the interactions it should attempt.

We believe that this is best addressed by requiring an agent to additionally pass a {\em do no harm} test in order to achieve a non-zero completion score. This test is a single binary predicate test which can be defined by the task designer in any way they choose. A simple choice  consists of a logical `and' of state tests on all non-task objects in the scene, checking that they have not been moved. The movement threshold for each object could be a simple rule such as requiring an IoU above a threshold between every object's starting and final poses.

More sophisticated `do no harm' tests could be tuned per object (since it may not matter if some non-task objects are moved) or could test other properties than pure object pose, such as the maximum acceleration or forces experienced by objects within the simulator.

\paragraph{General evaluation in terms of scene predicates.}
A more general formulation of completeness will not necessarily take the form of one binary threshold test per object or part, but would define a set of predicates $P_j(s,s*)$, each of which is a binary threshold test involving the individual final and target states of any number of objects. Overall completeness would be the percentage of these predicate tests which have been passed.

The most obvious use for more general predicates is to allow for tests of the relative poses of objects. For instance, if a saucer must be placed on a table, and a cup on the saucer, one predicate would set a threshold on the relative pose of the saucer and the table, and
another on the relative pose of the cup and saucer. As with individual object pose tests, these predicates could use just relative translation, or also angular measures. Each predicate test would be given a suitable specific success threshold.

Our focus in rearrangement is physical object manipulation, but defining completion in terms of a set of general logical predicate tests does in principle allow other non-physical binary tests to be part of an overall performance measure --- \eg whether a logical switch in the simulator has been touched to put it into the correct on/off state.

\paragraph{Evaluation programs.}
Evaluation of task completion should be carried out and reported automatically by the task simulator, and must therefore be implemented by an evaluation program. This program is passed the final and target state configurations once an agent has finished working, and performs the required threshold tests to determine the overall completion percentage.

The designers of agents should have a clear understanding of how evaluation is to take place, and therefore we propose that evaluation programs should be defined in a clear and public domain-specific language, which may eventually become a standard across different simulators.

Generally it is not appropriate that all of the parameters of the precise tests which make up a particular task should be published, because an important aspect of rearrangement agents is to be able to deduce the goal state of a task from the specification which could be provided in different forms as detailed in Section~\ref{sec:rearrangement}.
General information such as IoU thresholds could be made available, or it could be considered part of the task for the agent itself to use background semantic knowledge to deduce suitable thresholds for each type of object.

\paragraph{Emergent properties of simple evaluation metrics.}
Although the formulation of completion in terms of general predicate tests allows arbitrary flexibility in task specification, we would like to point out some of the interesting emergent properties of simple completion specifications. These arise due to the properties of a physically simulated environment with a limited scope and range of possibilities.

At first thought, it might seem that target locations specified in relative terms will quickly be needed. For instance, if a saucer must be placed on a table, a cup on the saucer, and a spoon in the cup, the target location of each object could be specified relative to the previous one. However, a well-defined threshold test in absolute coordinates could still capture the situation well, and in particular translation-only pose tests are often sufficient.
If the cup needs to achieve a height coordinate a few millimetres above the top of the table, in a task and environment where only a few objects are present the only way to achieve this might be to place it on the saucer, and if the centre of mass of the spoon needs to be somewhat higher still the only way to achieve this might be to place it in the cup. Similarly, if a key is to be placed in a lock, there will be no need to specify its target orientation, because within the physics simulator the only way to get the centre of mass of the key to the correct location will be at the orientation which fits into the lock.

When many objects must be packed into a tight space, but their order is unimportant, a simple threshold which is the same for all of the objects may work. For instance, if many books should be placed onto a bookshelf, each can be tested with a threshold based on the size of the whole shelf. The difficulty of placing many books in sequence will naturally increase as they must be fit into the smaller and smaller space that physics allows.
We believe that many rearrangement tasks where the placement of multiple objects involves physical coupling will share these emergent properties of incremental difficulty from simple tests. Another example would be building a tower from blocks, where each block is tested in terms of target height. The final blocks will only be able to pass their evaluation tests if the lower blocks have been placed into a stable tower, and the early blocks must be placed precisely if a tall tower is to be built.

\paragraph{Predicates based on natural language.}
It is worth noting that the need for an automatic evaluation program poses some challenges for rearrangement tasks specified with free-form natural language. Effectively this means that we need a program that can recognize whether the state achieved by an agent satisfies a natural language predicate. In many cases, such a program can be difficult to hand-craft. For example, the meaning of ``on'' may not be easily reducible to simple rules of geometric relations~-- ``a cup on a table'' would be very different from ``clock on the wall''. Thus even though simple evaluation programs are preferred, our formulation allows the evaluation program to take a more complex and less interpretable form, such as a neural network that has been trained to classify the ``on'' relationship.

\paragraph{Towards evaluation of real-world robotic agents.}
Despite our focus in this report on rearrangement in simulated environments, our general interests in Embodied AI mean that we have a long-term interest in whether the methods we propose are also applicable to real-world robotic agents. With regard to evaluation and the completion metric, this is certainly the case in principle, though automatic and accurate evaluation of real-world rearrangement presents huge technical challenges in anything but trivial tasks. Arbitrary objects would need to be tracked or instrumented in difficult situations of occlusion and contact. Building such systems will certainly be an area of important future research.

\subsection{Secondary Metrics}

Although we propose task completion as a unified primary metric, we believe that rearrangement simulators should also report additional measures of performance which are important in judging whether an agent has real-world value. In particular, we argue for metrics which measure efficiency, both of the performance of the agent in the simulated environment and of its computational requirements.

While multiple metrics can always be combined into single values via weighted addition or other formulae, this must be based on a choice of the relative importance of the different factors, and we believe that it is better to present the metrics separately so that potential users can make choices between agents based on their own criteria. An agent will usually have settings which allow performance metrics to be traded off against each other, and a sampling of these settings will produce a multi-objective performance manifold which can be displayed with a Pareto Front.

Useful secondary metrics include:

\begin{itemize}
    \item Simulation time taken for an agent to report completion and stop action: this is in units of the time defined within the simulation, or the number of simulation `ticks' required, with respect to which the agent program can take actions at a constant defined rate.
    \item Simulated energy required by the agent to take all of the actions within an episode. Again this is purely measured within the physics simulation, via integration of all of the virtual physical work done by the agent to move objects and its own body, and should be something that a good simulator is able to calculate. We believe that minimising simulated energy is ultimately a powerful and highly general metric that encompasses many aspects of efficiency and smoothness in an agent's actions.
    \item Computational measures of the agent program. How many FLOPS and how much memory does it need per `tick' of the simulation, or to complete a whole episode? One way to measure this could be if the agent program is running on the same machine as the simulator, and requesting tick updates of the simulator when it is ready to deal with them: at what factor of `real-time' in the simulator is the agent able to run?
\end{itemize}

These measures of efficiency are especially critical as the aim of building intelligent rearrangement agents is ultimately progress in real-world Spatial AI, where agents which must run on the limited embedded computation platforms in robots or other devices, although these platforms will surely be cloud-connected in general and some high-latency computation could be carried out remotely~\cite{Davison:ARXIV2018}.
Later more sophisticated computational measures should also encompass the degree to which an agent's computation and storage can be parallelized, distributed or layered in terms of latency.

\section{Experimental Testbeds}
\label{sec:testbeds}

\begin{figure*}
    \includegraphics[width=\textwidth]{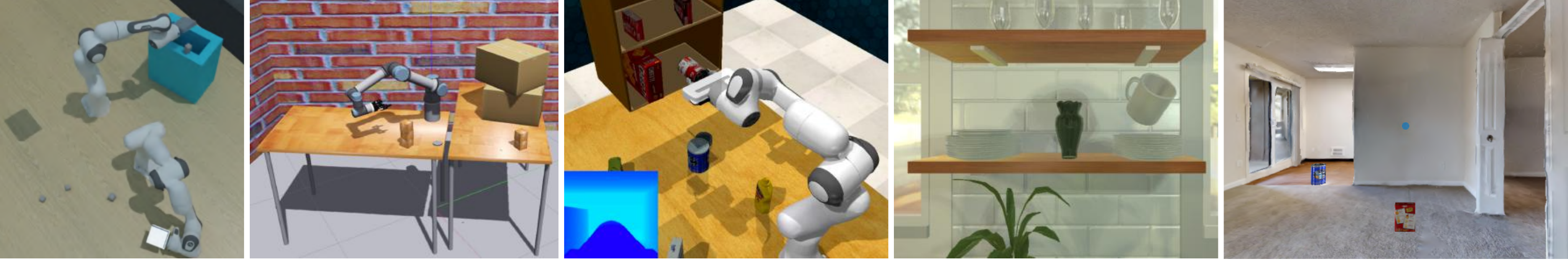}
    \caption{\textbf{Suite of experimental testbeds.} We contribute a set of experimental testbeds for rearrangement, spanning a spectrum of environment complexities, and agent navigation and manipulation capabilities. From left to right: T1: bimanual sweeping, T2: table organization, T3: storing groceries, T4: room rearrangement, T5: house cleanup.}
    \label{fig:testbeds}
\end{figure*}

Here we summarize a set of experimental testbeds for rearrangement that we contribute with this report (see \Cref{fig:testbeds}).
These testbeds span a spectrum of environment complexities, as well as agent navigation and manipulation capabilities.
We order the testbeds roughly by environment scale and navigational requirements (in addition to manipulating objects).

\paragraph{T1: bimanual sweeping.}
A pair of fixed-base robot arms sweep `trash' objects (simulated as small cubes) from the floor into a `trash bin' container.
Four third-person view camera and depth sensor pairs are positioned to observe the arms, and the position and velocity of the arm kinematic chain are also available.
Both of the arms have spatulas as end effectors to sweep cubes from the floor into the trash bin.
This scenario is implemented in SAPIEN~\cite{Xiang2020}. See \Cref{app:sweeping} for details.

\paragraph{T2: table organization.}
A fixed-base robot arm (Universal Robots UR5e) with a parallel-jaw gripper (Robotiq 2F-85) is tasked with rearranging a set of tabletop objects.
The objects start in a random configuration and need to be rearranged into a specific state on the table.
A stationary third-person and wrist-mounted camera and depth sensor are available for perception.
This task is also implemented in SAPIEN.
See \Cref{app:table_organization} for details.

\paragraph{T3: storing groceries.}
A fixed robotic arm manipulator (Franka Panda arm with Franka gripper) is tasked with picking up a randomly scattered set of grocery objects on a table and placing them into a constrained shelf space.
The target location for all of the objects is defined as the same volume above the shelf, rather than a specific goal for each object.
This leads to interesting emergent difficulty as objects will be increasingly difficult to place within the volume as the shelf becomes more occupied, and the best solutions require planning of the best order and placement for moving all of the objects.
The sensory suite includes color camera and depth sensors mounted on the wrist, hand and over-the-shoulder, as well as proprioceptive sensors including joint encoders and forces.
This scenario is instantiated in the RLBench task suite~\cite{James2020}.
See \Cref{app:groceries} for details.

\paragraph{T4: room rearrangement.}
This scenario involves rearranging randomly placed household objects in a room and changing the state of the objects, such as opening/closing a cabinet.
An example is shown in \Cref{fig:example}.
Identifying objects that have changed, inferring the state of the objects, planning a path for reaching and manipulating objects (\eg manipulating an object might require moving a blocking object) are among the challenges of performing this task.
The agent uses a `magic pointer' style manipulation.
The level of the difficulty of the task varies depending on the number of objects changed, their configuration in the scene and the complexity of actions required to recover the goal configuration.
The scenario is instantiated in AI2-THOR~\cite{Kolve2017}.
Refer to \Cref{app:thor} for the details of this rearrangement scenario.

\paragraph{T5: house cleanup.}
In this scenario, a mobile agent with a `magic pointer'-style manipulator is tasked with cleaning up a house.
The agent must find randomly placed household objects, pick them up, and move them to a different specified location.
The agent carries camera and depth sensors for perception.
As the task involves relocating objects between rooms, it involves longer-range navigation.
This scenario is implemented in Habitat~\cite{Savva2019} (see \Cref{app:habitat}).

\section{Why Rearrangement?}

\begin{figure*}
\includegraphics[width=\textwidth]{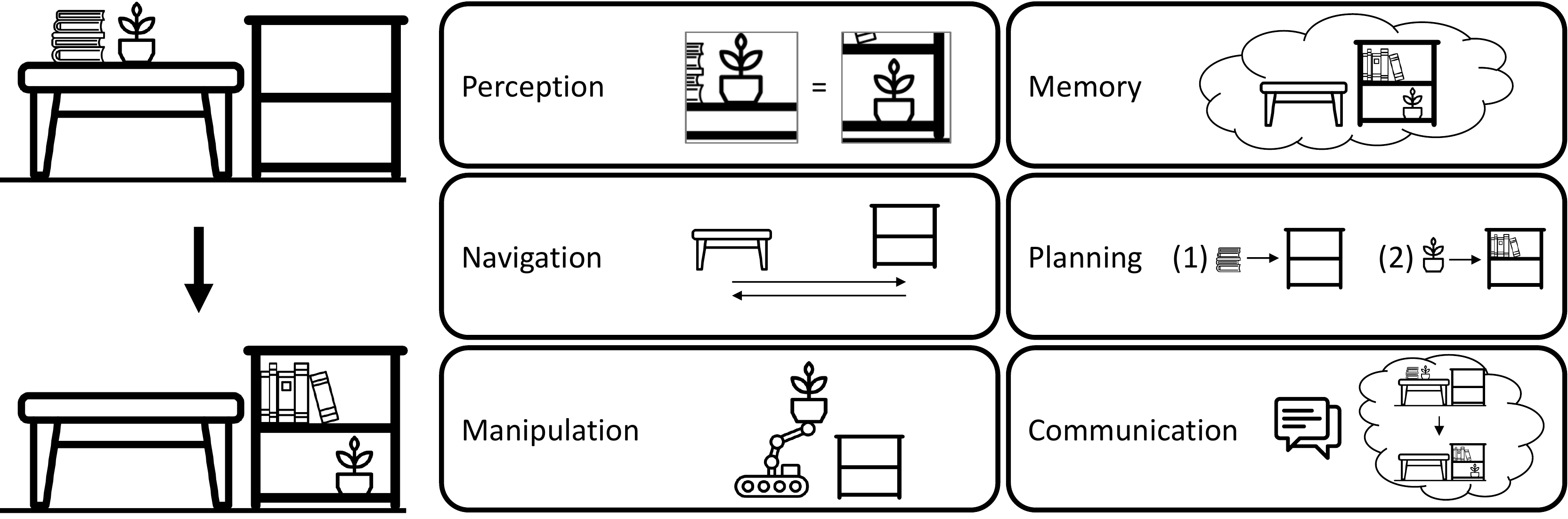}
\caption{\textbf{Agent abilities harnessed by rearrangement.} Illustration of the broad spectrum of abilities the rearrangement task can exercise.}
\label{fig:abilities}
\end{figure*}

The spectrum of possible rearrangement scenarios can be used to exercise and evaluate a broad set of agent abilities (see \Cref{fig:abilities}).
Different research communities may focus on analyzing and evaluating subsets of these abilities.
Here, we summarize several types of abilities that the rearrangement task can evaluate.

\paragraph{Perception.}
Evaluation of the agent's performance on subtasks such as object detection~\cite{liu2020deep}, rigid object state estimation and tracking~\cite{yilmaz2006object}, and localization~\cite{fuentes2015visual} of the agent relative to objects connects with a breadth of research problems in the vision community.
Perceptual abilities underlying these subtasks are important for an agent to identify object instances that can be moved, judge whether it is in a position that affords manipulation of an object, and to estimate the current state of the object (\eg is the cabinet door open or closed?).
Evaluating whether a target rearrangement has been achieved may also involve perception.
Perceptual abilities underlie both traditional vision tasks as well as emerging tasks such as prediction of object physical properties to enable interaction with the object (\eg ``is the bottle likely to break if it is dropped?'' or ``is the box light enough to be picked up or does it need to be pushed?''), or prediction of affordance heatmaps for objects~\cite{myers2015affordance}.
The degree to which these abilities are stressed is linked to the choice of agent sensors and environment complexity, and how sensory information can be used for action prediction.

\paragraph{Navigation.}
Controlling an agent to achieve locomotion is required when performing rearrangement within larger environments.
Consequently, rearrangement can be used to investigate agent navigation capabilities, a topic that is of interest to both the robotics and vision communities~\cite{bonin2008visual}.
The formalization of navigation tasks for Embodied AI agents has been described in more depth by Anderson et al.~\cite{Anderson2018}.
We can view the rearrangement task as a generalization of the navigation task as the agent is required to change its own state relative to the environment (i.e.\ to navigate), as well as to change the state of the environment (i.e.\ to manipulate).
Different choices of agent mobility and environment complexity lead to varying degrees of focus on collision-free navigation and path efficiency within a rearrangement task.

\paragraph{Manipulation.}
To successfully rearrange objects, an agent must grasp and manipulate the objects.
Grasping and manipulation are rich research areas within robotics~\cite{murray1994mathematical,mason2018toward}.
Different choices on the manipulation complexity axis lead to different levels of abstraction for manipulation.
High-level abstractions such as the ``magic pointer'' manipulator reduce the importance of manipulation and may be more appropriate for focusing on perception.
At a low level of abstraction, the rearrangement task can be used to investigate control policies for trajectory and grasp planning.
Different choices of the number and types of objects to be rearranged (\eg degree of instance variation in geometry and appearance), and varying the complexity of the environment (\eg fairly open spaces to cluttered environments) can control the focus on manipulation.

\paragraph{Memory.}
The rearrangement task also provides a testbed for investigating different types of memory and representation.
As rearrangement consists of a sequence of object manipulations and/or agent navigations, the ability to store and recall information about the environment and the self is important.
Investigating agent memory representations is a topic of great interest in the machine learning and Embodied AI communities~\cite{banino2018vector,cueva2018emergence,Davison:ARXIV2018}.
Simultaneous Localization and Mapping (SLAM) research in robotics has produced algorithms which can incrementally and consistently build 2D or 3D scene representations from raw data from cameras and other sensors.
These representations have gradually improved in geometric accuracy and completeness, and some SLAM systems now incorporate recognition and aim to build an explicit graph of interacting object instances which can be used to simulate and plan robot manipulation~\cite{Wada:etal:CVPR2020}.
An important research issue going forward is to what extent
challenging rearrangement tasks require agents with the ability to build this kind of explicit 3D scene representation designed into them, versus what can be achieved with the implicit representations used by
black-box agents trained via machine learning.

\paragraph{Planning.}
The rearrangement task requires planning a sequence of actions and reasoning about the pre-conditions and post-conditions for specific actions.
For example, to set a dinner table, the agent needs to: go to the kitchen, open the cabinet door, get plates, place the bigger plate on the table first, and then place the smaller plate on top of the bigger plate.
Doing the place settings for four people requires hierarchical decision making and sub-task ordering.
Agent architectures that can successfully plan for such structured decision-making scenarios are an emerging focus in Embodied AI.

\paragraph{Communication.}
The rearrangement task allows studying grounding of language to perception and action, a topic relevant to natural language researchers interested in Embodied AI~\cite{bisk2020experience}.
By specifying the ``LanguageGoal'' as either a series of instructions or a description of the final state of the environment, the task allows researchers to go beyond grounding nouns such as ``chair'' and ``table'', to spatial relations between objects and how they are expressed in different languages (``cup in cabinet'', ``painting on the wall''), to actions (``pick up'' vs ``put down'' vs ``open''), as well as higher-level concepts such as ``set the table for four'' and ``clear the table and wash the dishes''.
A successful agent will also need to retain ``common sense'' and implicit knowledge about target arrangements that may not be precisely specified in the language (\eg what objects and where they should be placed for ``set the table for four'').
The rearrangement task can also be used to study communication for coordination between multiple agents and emergent behaviors.

\paragraph{}
In summary, the rearrangement task provides a unified platform for investigating methods to endow agents with the above agent capabilities.
By making different choices on the complexity axes, we can emphasize specific capabilities (\eg more abstract agent control schemes to focus more on planning) or combinations of capabilities.

\section{Discussion}
\label{sec:discussion}

\paragraph{Comparison to existing benchmarks.}
The rearrangement task presents a significantly more difficult challenge than prior navigation-based benchmarks for Embodied AI.  Specifically, the degree of physical interaction with the environment is inherently much greater in rearrangement.  In fact, the mobile variant of the rearrangement task can be viewed as encompassing prior work on PointGoal navigation~\cite{Anderson2018}. Furthermore, rearrangement requires more precise reasoning about objects, including their physical properties (\eg placing one object inside another, pouring liquids, grouping similar items) and geometric constraints (\eg arranging books on a bookshelf).  As a result, the embodied agents developed for the rearrangement task will need to be far more sophisticated than those of today, helping to advance the development of more effective world representations, sequential decision-making algorithms, perception techniques, task-oriented grasping methodologies, and physics simulations.

\paragraph{Complex tasks and processes.}
Example tasks described in this document have largely consisted of rearrangement tasks in which the end goal is specified directly, without guidance as to the intermediate stages that the environment must go through to reach the goal.
However, many real-world tasks consist of complex processes that can be broken down into multiple sequential subtasks or subgoals.  For example, cleaning the living room may consist of multiple subgoals corresponding to different parts of the room, or putting away groceries may be broken down by items that belong in the pantry, refrigerator, freezer, and counter.  In the context of complex tasks, rearrangement can be viewed as sequentially addressing each individual subgoal.  Many avenues for future work exist in this area, particularly in ordering subgoals and resolving dependencies between subgoals.

\paragraph{Application to physical robotic systems.}
The ultimate goal of Embodied AI is to develop systems that perceive and act in physical environments –- i.e.\ physical robots in physical worlds.  We believe that the simulation environments, tasks, and evaluation procedures proposed in this document directly aid this goal.  Development in simulation presents a number of important benefits, including access to orders of magnitude more data, more precise evaluation techniques, and more efficient experimental procedures, which will all fuel more rapid research progress.  However, this work must take into account a number of potential downsides to simulation-based development -- mainly the potential development of techniques whose performance in simulation does not transfer to real-world domains.  To address the potential disconnect between simulated and real-world performance, we propose that physical robot variants of the rearrangement tasks be also developed in parallel.  Robust evaluation of performance (i.e.\ automated techniques for assessing the difference between the current state and the goal) remains the most challenging component of such an implementation.  To aid progress in this area, early variants of the rearrangement task could be based on either simple table-top domains, or on instrumented environments.  Evaluation could be based on standard object datasets, similar to the YCB dataset \cite{calli2017yale}.  While more general and scalable evaluation techniques will be needed in the long term, in the short term such experimental domains can be used to ensure strong correlation between simulated and real-world experiments, such that progress in simulation effectively translates to, or is predictive of, performance on real-world systems.

\paragraph{Simulation fidelity.}
The fidelity of the simulator in terms of both the underlying physics and simulated sensing modalities will significantly influence task complexity. When investigating the rearrangement task in simulation, with the aim to transfer a learned model to the real world, the gap between simulation and reality will impact the additional effort to achieve successful transfer. Actuation and sensing noise can be simulated to better approximate real-world robots, adding complexity to the task. We believe that current simulation platforms are well-equipped to simulate rigid body dynamics, optionally with noise models on important physical parameters. Efficient and accurate simulation of deformable objects and phenomena such as fluids remains challenging and is a good direction for future work.

\paragraph{Future extensions.}
In this report, we propose a number of concrete instantiations of the rearrangement task. Importantly however, the proposed task leaves room for further extensions that would increase complexity and more comprehensively exercise the intelligence of the agent.  Below, we discuss several extensions to the rearrangement task that are beyond the immediate horizon but that we believe are promising directions for future work.

\begin{itemize}
    \item \textit{Deformable objects:} The ability to manipulate and rearrange deformable objects (\eg clothing, towels, curtains, or bed sheets) has many practical applications in everyday environments. However, deformable objects introduce many challenges, including simulating deformations, defining appropriate goal specifications, and evaluating the similarity between the end result and the goal. These challenges are significant, and thus preclude the inclusion of deformable objects in our initial formalization and instantiation of Rearrangement. This is an obvious practical extension for future years.

    \item \textit{Transformation of object state:}
    Many everyday tasks involve the transformation of object state.  Examples include pouring water into a glass, chopping vegetables, heating a skillet, washing a dish, turning on an oven, or whisking an egg. These state changes are fundamentally different from rigid-body transformations.  They i) require the incorporation of deeper world knowledge, ii) rely on causal reasoning methods to interpret, iii) involve the execution of complex actions beyond pick-and-place, and iv) necessitate separate procedures to specify and evaluate such scenarios. As a result, we do not address such tasks within the scope of the presented formulation, although the predicate-based evaluation that we propose allows for natural future extensions of the rearrangement task in this important direction.

    \item \textit{Multi-agent rearrangement:} In this report, we focused on single agents performing the rearrangement task. We believe that extending the rearrangement task to multiple agents is natural and interesting. There are three types of agents that may be present: cooperative agents, adversarial agents, and non-participating agents. Cooperative agents can coordinate to achieve the task more efficiently than a single agent.  Certain scenarios may require two agents to cooperate to accomplish a subgoal, such as moving a heavy sofa. Adversarial agents can actively prevent the agent performing rearrangement from achieving the task. These agents may either provide inaccurate or false information, or they may actively disrupt the agent or the environment. Non-participating agents are present but do not actively work towards or against the rearrangement-performing agent(s). The number of agents of each type may vary. There is also a spectrum of communication mechanisms that may be available to agents.

    \item \textit{Interactions with human users:}
    All examples in our work are ultimately driven by the desire to develop systems that are able to assist users in a wide variety of everyday tasks. Toward this goal, future work should examine human interactions with rearrangement systems.  Example topics include intuitive real-time interactions, dialog-based interfaces, and active learning from human input, among others.

\end{itemize}

We leave these and other interesting extensions of rearrangement as promising directions for future work.


\section*{Acknowledgments}

We thank Ankur Handa, Camillo J. Taylor, Deepak Pathak, Dieter Fox, Dmitry Berenson, George Konidaris, Jana Kosecka, Josh Tenenbaum, Ken Goldberg, Kostas Daniilidis, Kristen Grauman, Leslie Kaelbling, Lucas Manuelli, Matthew T. Mason, Niko S{\"u}nderhauf, Oliver Brock, Pete Florence, Peter Corke, Pieter Abbeel, Raia Hadsell, Richard Newcombe, Russ Tedrake, Saurabh Gupta, Shubham Tulsiani, Siddhartha Srinivasa, Stefanie Tellex, Tomás Lozano-Pérez, and Vincent Vanhoucke for their feedback on a draft of this report. We thank Joseph Lim for participating in early discussions. We also thank the AI2-THOR, Habitat, RL-Bench, and SAPIEN teams for releasing the experimental testbeds described in this report.

\balance

{\small
\bibliographystyle{ieee}
\bibliography{rearrangement}
}

\clearpage

\newpage
\appendix

\section{Experimental Testbed Details}
\label{sec:testbed-details}

To support research in the immediate term, we release a number of rearrangement scenarios within a set of existing simulators.
We leverage AI2-THOR~\cite{Kolve2017}, Habitat~\cite{Savva2019}, RL-Bench~\cite{James2020}, and SAPIEN~\cite{Xiang2020}, but any simulator that supports the rearrangement task can be used in the future. A summary of the task specifications is provided in Table~\ref{tab:task_spec}.

\begin{table*}[tp]
\footnotesize
\setlength{\tabcolsep}{4pt}
	\centering
	\begin{tabular}{l c c c c c}
	    \toprule
	      & T1 & T2 & T3 & T4 & T5\tabularnewline
	      \midrule
	      \textbf{Goal specification} & PredicateGoal & GeometricGoal & GeometricGoal & ExperienceGoal & GeometricGoal \tabularnewline
          \midrule
	      \multirow{2}{*}{\textbf{Manipulation type}} & Spate  & Parallel Gripper & Parallel Gripper & Magic pointer & Magic pointer \tabularnewline
	      & (Coulomb contact model) & (Coulomb contact model) &&&
	      \tabularnewline
          \midrule
          \multirow{2}{*}{
	      \textbf{Perception}} & RGBD, joint kinematics & RGBD, joint kinematics & RGBD, joint & RGBD, Haptic & RGBD, GPS+Compass \tabularnewline &&& kinematics/torques &&
	      \tabularnewline
          \midrule
          \multirow{2}{*}{\textbf{Action space}} & Manipulation, & Manipulation by  & Manipulation, & Navigation, Manipulation & Navigation, Grab/Release \tabularnewline
                                & Grab/Release & ROS controllers & Gripper open/close & State change & \tabularnewline
	    \bottomrule
	\end{tabular}
	\vspace{1mm}
	\caption{\textbf{Rearrangement tasks summary.} A summary of the specifications for each of the experimental testbeds is provided. Detailed descriptions of the tasks are provided in Appendix~\ref{sec:testbed-details}.}
	\label{tab:task_spec}
\end{table*}

\subsection{Bimanual Sweeping in SAPIEN}
\label{app:sweeping}

\begin{figure}[ht!]
    \centering
    \includegraphics[width=0.8\linewidth]{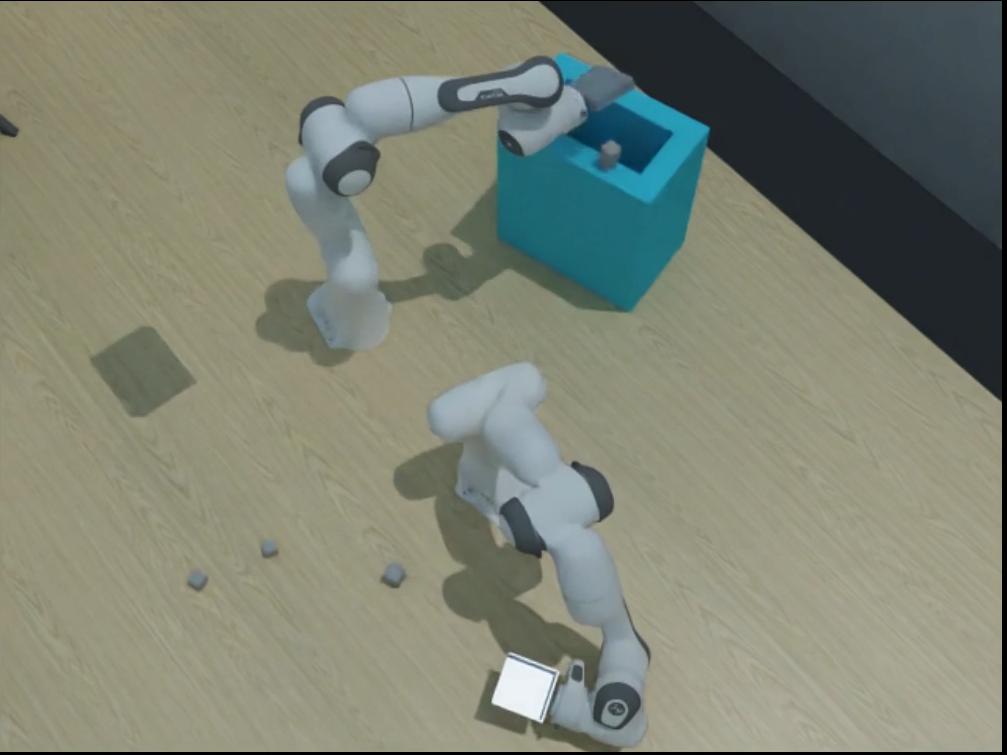}
    \caption{Bimanual sweeping task.}
    \label{fig:bimanual}
\end{figure}

This scenario is instantiated in SAPIEN~\cite{Xiang2020}, orignially as a final project in a robot learning course at UCSD. Participants are expected to control two robot arms to collaboratively pick up boxes randomly placed on the table and place them into the target bin efficiently. The task setup has been made available at \url{https://github.com/haosulab/CSE291-G00-SP20}.

\paragraph{Simulation Speed.} SAPIEN allows using two renderers for this task: 1) a rasterization-based Vulkan engine to render scenes at 200-300 FPS (30-60 FPS if GPU-CPU data transfer is used), and 2) the ray-tracing based OptiX engine to render scenes with near-photorealistic appearance at 1 FPS. Physics simulation is supported by the PhysX engine, which performs rigid-body simulation at about 5000 Hz.

\paragraph{Scenes and Objects.}
The scene contains 2 robot arms, 10 boxes, and a trash bin, all of which are placed on a table as shown in Figure~\ref{fig:bimanual}. The positions of two robot arm bases are constant, while configurations of boxes and the trash bin (size, position, thickness) are generated randomly at the reset of the environment. Once generated, the trash bin is fixed on the table.

\paragraph{Episodes.}
Each episode ends when it reaches the predefined maximum time steps or when the agent returns \textit{False}. It is noted that the environment will not end itself even all boxes have been placed correctly. Agents are expected to return \textit{False} if they believe the task gets accomplished or stuck.

\paragraph{Embodiment.}
The robot arms used in this task are simulated Panda 7-axes robot arms by Franka. We replace the end effector with a spade. The arms are fully physically simulated based on physics simulation within SAPIEN. The control interfaces and action space will be introduced later.

\paragraph{Sensors.}
Visual observations from fixed RGBD cameras at 4 different viewpoints (front, left, right, top) are provided with known intrinsic and extrinsic parameters. We optionally provide ground-truth object segmentation to avoid introducing a vision challenge. Next, we provide the joint positions and velocities of the robot arm as robot state observation.

\paragraph{Action Space.}
Two control interfaces are available which are common in real robot control: 1. joint position and velocity control based on PD controllers with tunable drive parameters; 2. direct joint torque control. Convenient physical properties of the robot arm, including generalized inertia matrices, kinematic Jacobians, and forward/inverse dynamics, are also available to participants.

\paragraph{Metric.}
The objective is to place as many boxes as possible into the target bin within fixed time steps. Following the metric defined in Section~\ref{sec:evaluation}, we introduce two metrics: \textit{success rate} and \textit{efficiency}.  \textit{Success rate} is the fraction of the number of boxes correctly placed among the number of boxes observed in all the scenes. \textit{Efficiency} is the average number of boxes correctly placed per minute. Given a fixed time budget, \textit{efficiency} is equivalent to the total number of boxes correctly placed.

\subsection{Cloud Robot Table Organization Challenge (SAPIEN \& Real robots)}
\label{app:table_organization}
\begin{figure}[ht!]
    \centering
    \includegraphics[width=\linewidth]{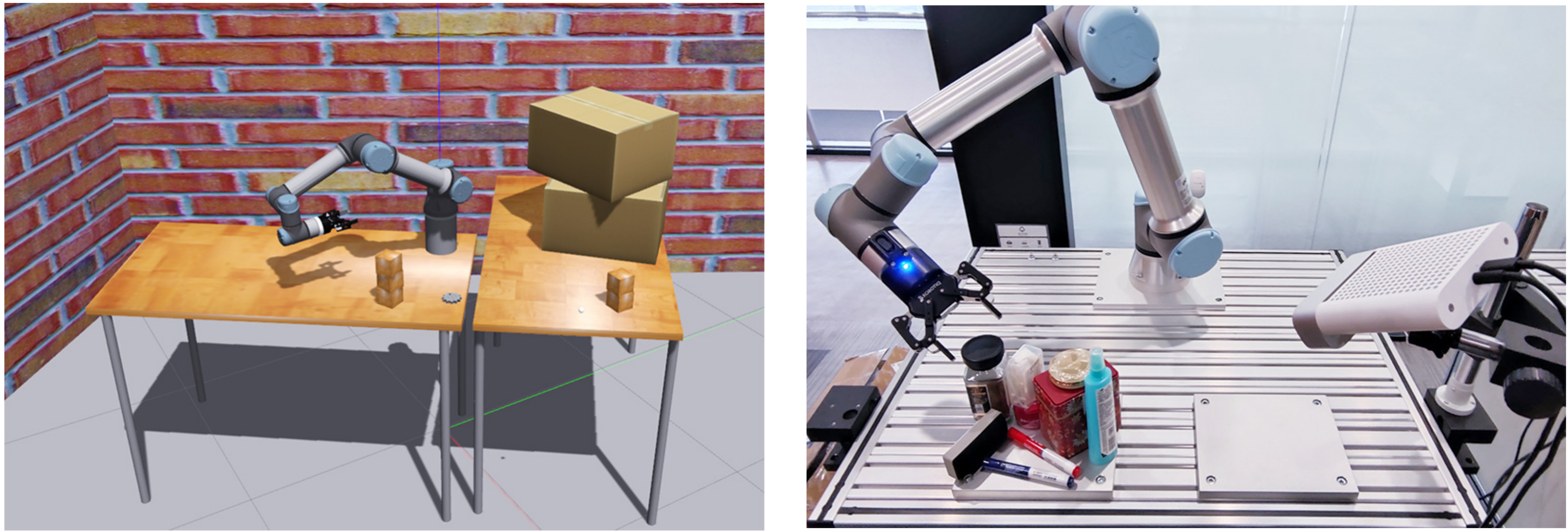}
    \caption{The IROS 2020 table organization challenge setup in simulator and the real world.}
    \label{fig:sim_and_real}
\end{figure}
In conjunction with IROS 2020, Su et al.\ hold a challenge which focuses on the task of table organization (see Figure~\ref{fig:sim_and_real}). The data and software of the challenge can be downloaded from \url{http://ocrtoc.org/}.

The competition contains two stages: simulation and real robot stage. Each stage begins with a trial period and ends with a contest. In the trial period participants can get familiar with the working environment and get prepared for the contest.

\paragraph{Stage1: Simulation.}
In the trial period of the simulation stage, participants can download the simulation package provided by the organizer, which comes with 1100 randomly generated initial/target scene pairs. Contestants are allowed to choose either SAPIEN~\cite{Xiang2020} or Gazebo as the simulation platform. They can try out the scenarios and get familiar with the software environment. In the simulation contest, participants need to upload their solution to the competition platform where their solution will be evaluated on 300 additional scene pairs.

\paragraph{Stage2: Real robot.}
In the trial period participants can test their solution on real robots.
The robot is controlled by a PC with GPU, on which the solution of participants will run.
In the contest period participants need to solve several tasks. In the end they will be ranked according to the metrics introduced below.

The setup of the real robot contains the following hardware:
\begin{itemize}
    \item A stationary camera: Kinect DK camera,
    \item A wrist-mounted camera: RealSense D435i,
    \item A manipulator: UR5e from Universal Robots,
    \item An end-effector: a parallel-jaw gripper 2F-85 from Robotiq,
    \item A PC as the computing device: CPU Intel Xeon E-2246G, Memory 32GB DDR4, GPU Nvidia Geforce RTX2080 with 8GB memory.
\end{itemize}

\paragraph{Simulation Speed.} The simulation speed considerations are the same as in experimental testbed T1 (Section~\ref{app:sweeping}).

\paragraph{Scenes.}
There are five different difficulty levels of scenes with regard to object geometry complexity and clutterness:

\begin{itemize}
    \item Level 1: 5 objects with simple geometry (box, can, etc.). For target configuration, all objects are placed on the table with no heap.

    \item Level 2: 5-10 objects with simple and complex geometry. For target configuration, all objects are placed on the table with no heap.

    \item Level 3: 10 objects with complex geometry. For target configuration, there are relative position specifications, e.g. a cup on the saucer, stacked boxes.

    \item Level 4: 10 objects with complex geometry, and 5 disturbing objects that are not scored in the target configuration. For target configuration, there are relative position specifications.

    \item Level 5: 10 objects with complex geometry, and 10 disturbing objects. For target configuration, there are complex and hierarchical relative position specifications.
\end{itemize}

\paragraph{Objects.}
We include various categories of daily life objects in the task. In general, object models (triangle mesh models with texture) will be given for task solving. However, there are different difficulty levels:

\begin{itemize}
\item Known objects with precise models: For these objects, precise object mesh models are provided (already during the trial period). These objects will be used in the trial period so that contestants can test their solution during the trial. In addition, these objects will also be used in tasks (low- to mid-level) of the contest.

\item Novel objects with precise models: For these objects, precise object mesh models are provided (in the contest). However, these objects will not be disclosed to contestants before the contest. They will be used in selected tasks (mid- to high-level) in the contest.

\item Novel objects with imprecise models: For these objects, only object models with similar geometry from the same semantic category are provided. These instance-level novel objects will only be used in tasks of high difficulty level in the contest.
\end{itemize}

\paragraph{Sensors.}
As described above, there are two RGBD cameras, one mounted to the robot wrist, the other fixed at the top left.
Also, the joint positions and velocities of the robot arm are provided as robot state observation.

\paragraph{Action space.}
The ROS Control \cite{chitta2017ros_control} interface is provided to command the robot in both simulation stage and real robot stage. Users can use any standard controllers defined in the protocol.
To control the robot arm, user can utilize \textit{joint\_trajectory\_controller} with joint-space trajectories on a group of joints. Trajectories are specified as a set of waypoints to be reached at specific time instants, which the controller attempts to execute as long as the mechanism allows. Waypoints consist of positions, and optionally velocities and accelerations.

\begin{figure}[ht!]
    \centering
    \includegraphics[width=\linewidth]{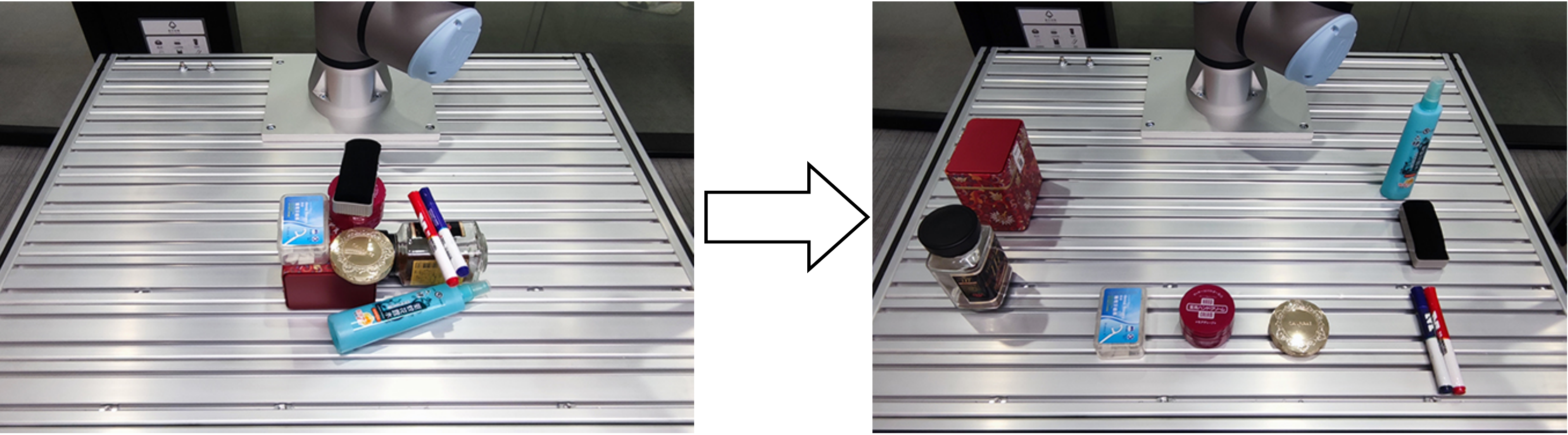}
    \caption{The initial and target arrangement of objects in the open cloud table organization challenge.}
    \label{fig:ocrtoc_states}
\end{figure}

\paragraph{Task.} We define a pool of tasks, which combines typical working scenarios in the context of service robots. All tasks will be mixed together in a task pool that is organized by the difficulty level of the task. All the tasks must be solved autonomously without any human intervention. In Figure~\ref{fig:ocrtoc_states}, an example task is provided to illustrate the key idea.

\paragraph{Metrics.} Following the recommendations in Section~\ref{sec:evaluation}, we use task completion as the primary metric, and also provide informative secondary metrics. Specifically, we measure how many objects were correctly organized. For each object in the target configuration, a distance error is calculated based on the difference between the actual pose and the target pose. There is a threshold for the distance error for each object. For each episode, the number of correctly rearranged objects is aggregated to compute the task completion metric.
Additionally, we also calculate the average distance error (in centimeters) and the execution time for each team.

\subsection{Storing Groceries in RLBench}
\label{app:groceries}

\begin{figure}[ht!]
    \centering
    \includegraphics[width=1.0\linewidth]{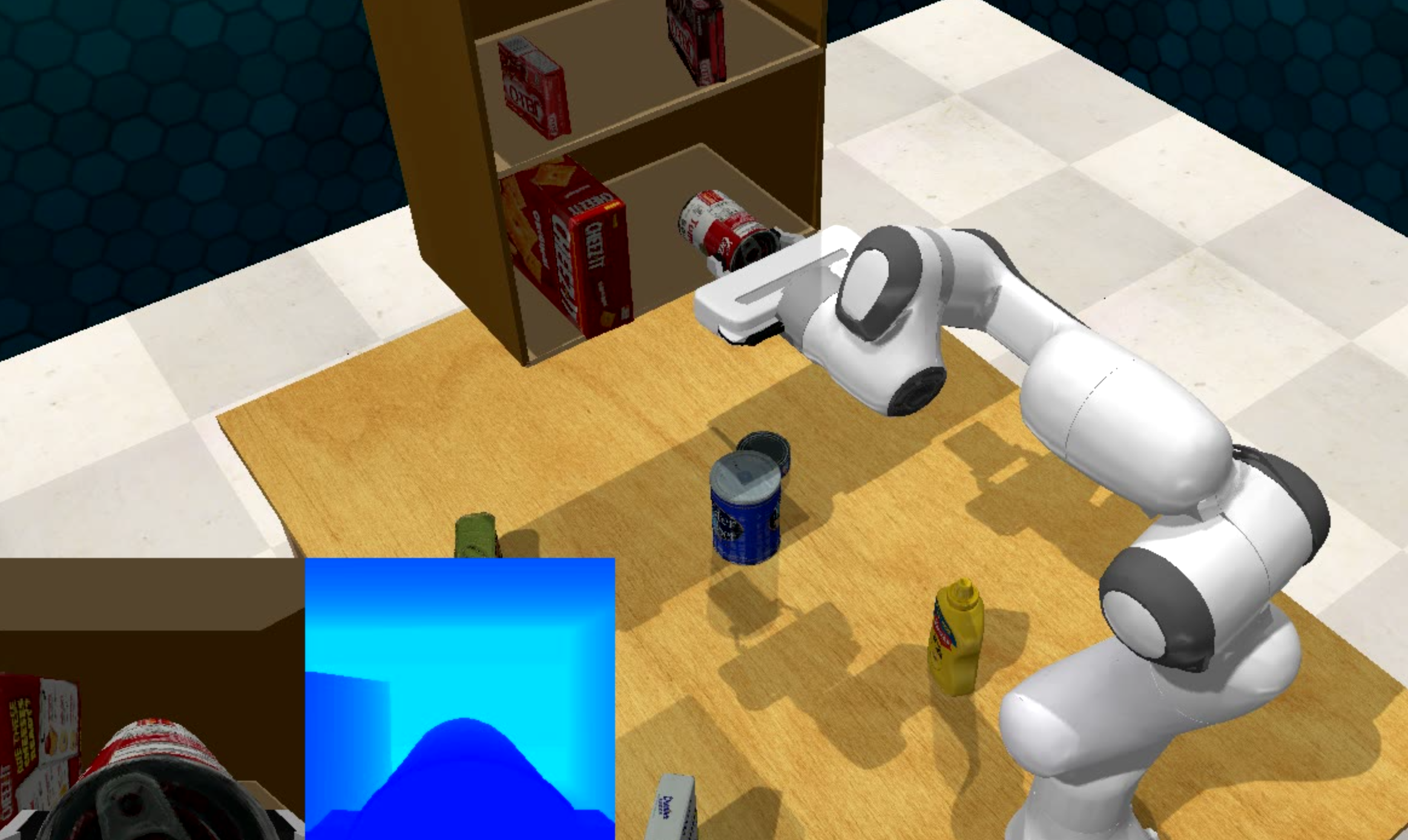}
    \caption{Storing groceries in RLBench.}
    \label{fig:groceries}
\end{figure}

This scenario (see \Cref{fig:groceries}) is instantiated in the RLBench task suite from the Dyson Robotics Lab at Imperial College~\cite{James2020}, available from \url{https://github.com/stepjam/RLBench}.
Having installed RLBench, see \url{https://sites.google.com/view/rlbench-rearrangement} for details and code for setting up the Storing Groceries Rearrangement Task.

\paragraph{Introduction to RLBench.}

RLBench is a robot simulation environment which offers over 100 different tasks for training and testing embodied agents. The emphasis is on variety of realistic tasks  that could be undertaken by a single fixed robot arm, using many different types of objects, and RLBench's original inspiration was as a testbed for meta-learning: to what extent are abilities learned to solve one task useful in other tasks, and is there structure and hierarchy among a large number of tasks whose relationship is not immediately obvious?

Each task is human-designed using intuitive tools provided within RLBench. A special feature of RLBench is that random variations of each task can be automatically generated, such as the starting locations of objects, and that for any variation an automatic demonstration can be generated, where the robot uses precise state information and motion planning to solve the task. These demonstrations can be used to seed reinforcement learning algorithmss.

Many of the tasks currently built into RLBench are rearrangement tasks, including such things as setting up a checkers board, loading objects into a dishwasher, emptying objects from a bin, taking a tray out of an oven, or making an ordered tower from blocks. We have selected the task of putting grocery objects onto a shelf to highlight here because it involves putting variety of interestingly-shaped objects into a constrained space, requiring high level planning as well as precision perception and manipulation skills.

RLBench is built on top of the CoppeliaSim robot simulator previously known as V-REP, available from \url{https://www.coppeliarobotics.com}.

\paragraph{Scene.}

The scene consists of 7 objects randomly placed on a table surface within reach of the robot arm, and a box shelf space in front of the robot onto which the objects must be placed.

\paragraph{Objects.}

The objects are accurately modeled grocery objects from the YCB dataset~\cite{calli2015ycb}.

\paragraph{Embodiment.}

RLBench features a simulated Franka Panda arm with a two-finger Franka gripper. The arm and objects are fully physically simulated, based on physics simulation within CoppeliaSim via Bullet. (Note that CoppeliaSim also offers four other physics engines, and these can be easily selected between.)

\paragraph{Action space.}

Various control modes for the robot arm are available which are familiar from continuous control of real-world robots, including direct velocity or torque control of the arm joints, or end-effector action modes where the agent can directly control the pose or velocity of the robot gripper.

\paragraph{Sensors specification.}

The sensory suite is broad, and includes proprioceptive force/torque sensing on all arm joints, and multiple color and depth cameras. Two cameras are mounted statically over the shoulder, and two others are located on the wrist and gripper. These function throughout operation, even when the robot has grasped an object which may cause significant occlusion.

\paragraph{Evaluation.}

Evaluation of the final state of each object is via a simple threshold on its translational pose, testing whether it is within the volume of the box shelf. The state tests are implemented in RLBench via simulated `proximity sensors' in its specification language, the sensing space of which can be easily visualised which is very helpful when designing tasks.

\subsection{Room rearrangement in AI2-THOR}
\label{app:thor}

\begin{figure}
    \centering
    \includegraphics[width=\columnwidth]{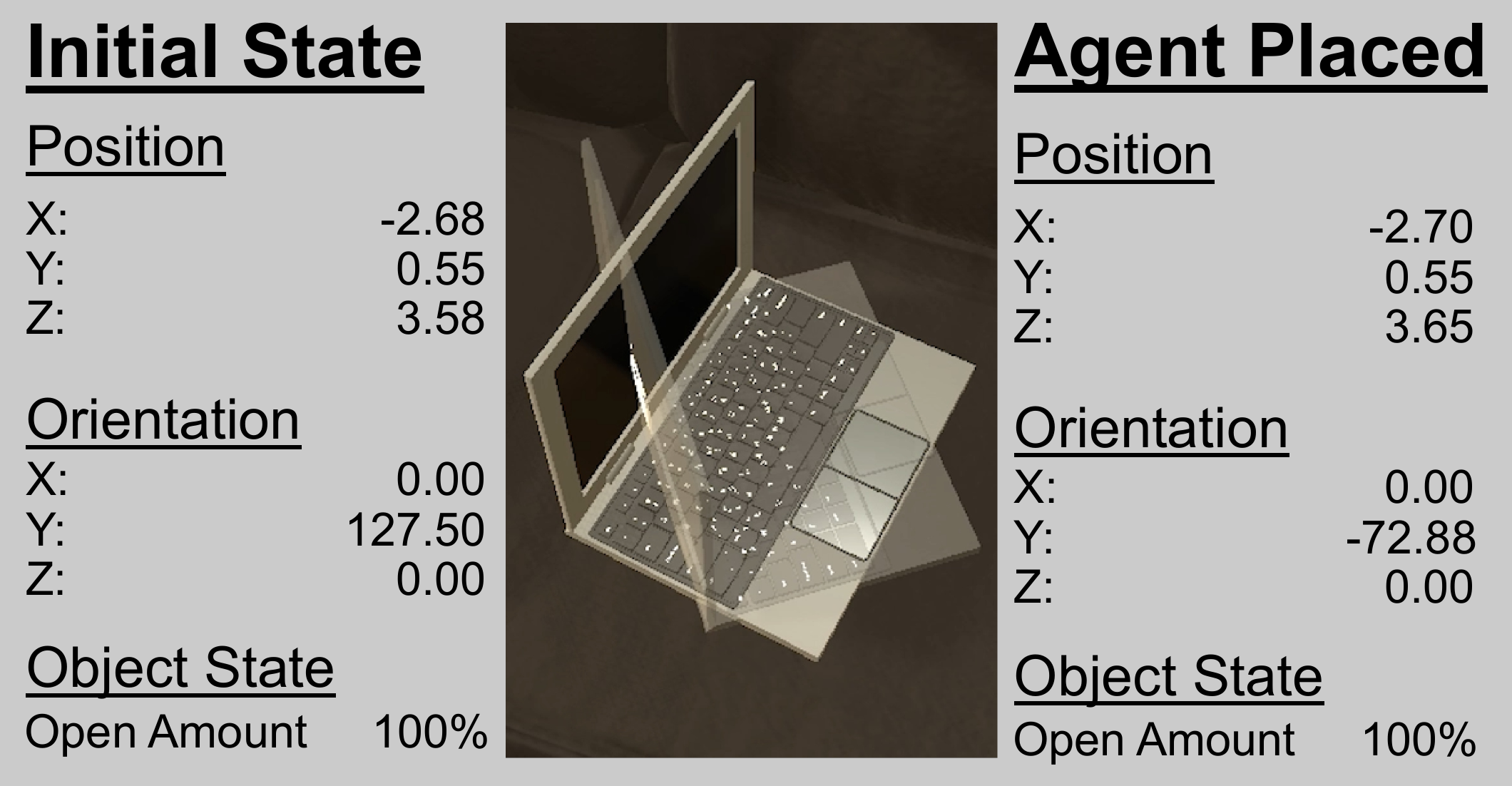}
    \caption{\textbf{Room rearrangement evaluation in AI2-THOR.} The pose of the laptop and its open/close state are used for computing the evaluation metric.}
    \label{fig:thor_eval}
\end{figure}

This scenario is instantiated in AI2-THOR~\cite{Kolve2017}. It involves rearranging randomly placed household objects in a room. More specifically, the scene has an initial configuration. We make changes to the scene by placing $N$ objects at different locations or changing their state (only open/close state is considered for this version). The task of the agent is to recover the initial configuration of the scene. The agent is allowed to navigate within the scene with the initial configuration and collect data for 1000 steps. This task specification falls under the category of ExperienceGoal described in Section~\ref{sec:rearrangement}.

\paragraph{Scenes.} We use the scenes of iTHOR for version 0.1 of the dataset. It includes 120 rooms across four categories (bathroom, bedroom, kitchen, and livingroom). Each category includes 30 rooms. Following the standard practice for AI2-THOR, we use the first 20 scenes in each category for training, the next 5 scenes for validation and the last 5 scenes for test. The agent is allowed to navigate and interact with the test scenes at their initial configuration. However, no metadata (object positions in 3D, segmentation masks, etc.) is available at test time. The metadata is available only for training and validation scenes.

\paragraph{Objects.} There are 125 object categories in AI2-THOR. Version 0.1 of the rearrangement dataset includes 53 categories. These include mostly categories that can be moved around easily (for example, small objects such as mug that can be moved to several different locations).

\paragraph{Dataset.} Version 0.1 of the rearrangement dataset includes 4000, 1000, 1000 scenarios for training, validation, and test, respectively. There are scenarios with different levels of difficulty. The number of objects whose state has changed varies, but we limit the maximum number to 5. Some scenarios involve only moving objects to a different location, while some involve changing the state (e.g., open/close a fridge). The dataset can be accessed at the following link: \url{https://ai2thor.allenai.org/rearrangement} .

\paragraph{Embodiment.} The collision geometry for the agent is a capsule with height 1.8m and radius 0.2m. The agent has a virtual arm, which is defined by a radius around the center of the camera i.e. the agent can move and manipulate objects anywhere within that radius. The virtual arm can go anywhere within the camera's frustum and within the agent's interaction distance (the default value is 1.5m). We consider a single agent for this version of the task specification.

\paragraph{Action space.} There are two types of actions we consider for the rearrangement task: Navigation and Manipulation actions. Navigation actions include: Move Forward, Turn Right ($\theta$ degrees), Turn Left ($\theta$ degrees), Look Up ($\theta$ degrees), Look Down ($\theta$ degrees). For simplicity, we assume the agent moves on a grid of adjustable size, but the simulator supports continuous and noisy movements as well. Manipulation actions include: Open/Close (point on the image), Pick Up (point on the image), Drop, Move Hand (to a relative x, y, z coordinate if allowed), Rotate Hand ($\theta$ degrees around x, y, or z axes) and Apply Force (point on the image, magnitude, direction). For actions that require a point on the image, the agent specifies a point. We trace a 3D ray from the camera center to that point. We apply that action to the first object that the ray hits. The object should be within a threshold of distance so the action succeeds.

\paragraph{Sensors specification.} We use three types of sensors for this version of the dataset: RGB, depth, haptic feedback. The haptic feedback indicates whether the virtual arm of the agent has touched an object or not. If the arm touches an object, the arm length will be returned to the agent. During training and validation, the simulator also returns the type of the touched object. The category information must not be used during test.

\paragraph{Metric.} Following the metric defined in Section~\ref{sec:evaluation}, we compute the average of the percentage of the satisfied predicated for each scenario. The predicate that we consider for this task is a conjunction of two propositions: (a) The IOU of the bounding boxes for the agent placement and the groundtruth placement of the object should be more than 50\%. (b) The object's `open/close' state should be within 20\% of the groundtruth. For example, if the fridge is closed, the task is considered successful if the fridge is at most 20\% open. Figure~\ref{fig:thor_eval} shows an example of agent and groundtruth placements along with the parameters used for the metric. Note that the episode will be considered unsuccessful if the agent changes objects that are not changed at the initial and the goal states of the scene.

\subsection{House Cleanup in Habitat}
\label{app:habitat}
In this scenario, the agent is spawned randomly in a house and is asked to find a small set of objects scattered around the house and place them in their desired final position as efficiently as possible.
In the following, we will describe the agent's observation space, action space, dataset and evaluation metrics in more detail. This scenario is instantiated in AI Habitat~\cite{Savva2019}, with code and data available at the following link: \url{https://dexter1691.github.io/projects/rearrangement/}.

\paragraph{Scenes.}
We use a manually-selected subset of 55 photo-realistic scans (35 for training, 10 for validation, 10 for testing)
of indoor environments from the Gibson dataset~\cite{Xia2018}.
These scenes are uncluttered `empty' apartments/houses, \ie they do not contain any furniture as part of the scanned mesh.
Scanned object meshes are programmatically inserted into these scenes to create scenarios.
This combination of empty houses and inserted objects allows for
controlled generation of training and testing episodes. Moreover, this setup ensures that all
objects in the house are interactive.
Notice that if we had used non-empty houses from Gibson, the objects included in the house scan would be non-interactive (since Gibson scans are static meshes).
This would result in an artificial separation between static baked objects and dynamic inserted objects.

\begin{figure}[t]
    \centering
    \includegraphics[width=1.0\linewidth]{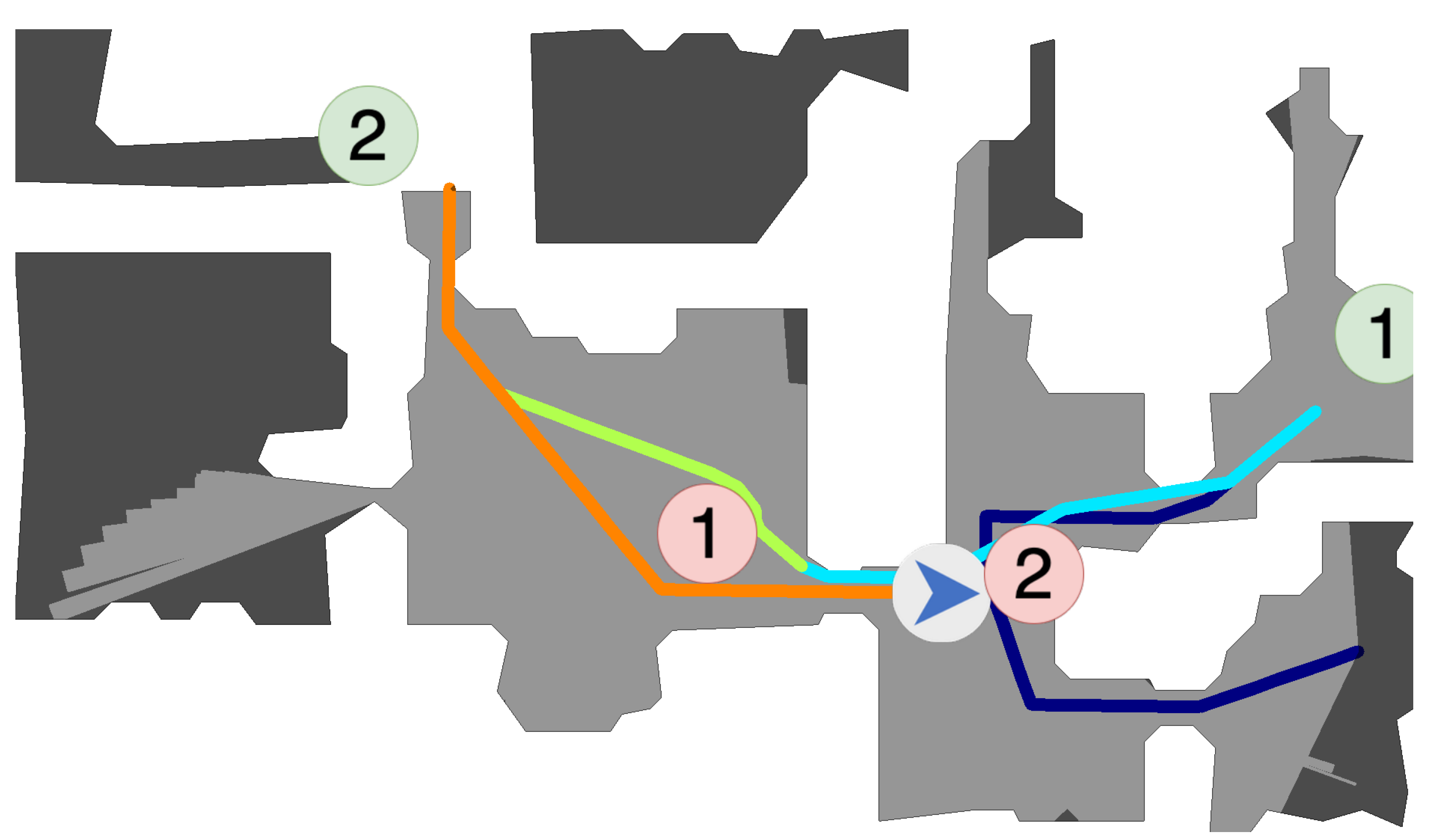}
    \caption{Rearrangement task in Habitat: Top-down visualization of a single rearrangement episode.
    Green circles denote the starting locations of two objects (1 and 2) and red circles denote the goal positions for the two objects.
    White circle with blue arrow denotes current agent location and pose. The colored lines indicates the path taken by an agent to solve this episode.}
    \label{fig:teaser}
\end{figure}

\paragraph{Objects.}
We use object scans from the YCB Dataset~\cite{calli2015ycb}.
These objects are small enough
they can pass through doors and hallways within the house.

\paragraph{Episodes.}
Each episode requires the agent to rearrange 2-5 objects.
The episode definition follows the GeometricGoal specification and consists of the scan name, spawn location and rotation of the agent, initial object location, rotation, and type in the environment. Finally, for each goal object, the episode defines initial and desired position of the center of mass.

\paragraph{Embodiment.} The agent is a virtual Locobot~\cite{locobot}. The simulated agent’s base-radius is 0.61m and the height is 0.175m which matches the LoCoBot dimensions.

\paragraph{Sensors.} Similar to the PointGoal navigation task in Habitat~\cite{Anderson2018,Savva2019}, the agent is equipped with an RGB-D camera placed at the height of 1.5m from the center of the agent's base and is looking in the `forward' direction. The sensor has a resolution of 256x256 pixels and a 90 degree field of view. To mimic the depth camera's limitations, we clip simulated depth sensing to 10m. The agent is also equipped with a \texttt{GPS+Compass} sensor, providing agent location (x, y, z) and heading (azimuth angle) in an episodic coordinate system defined by agent’s spawn location (origin) and heading ($0\degree$).

\begin{figure}[h]
    \centering
    \includegraphics[width=1.0\linewidth]{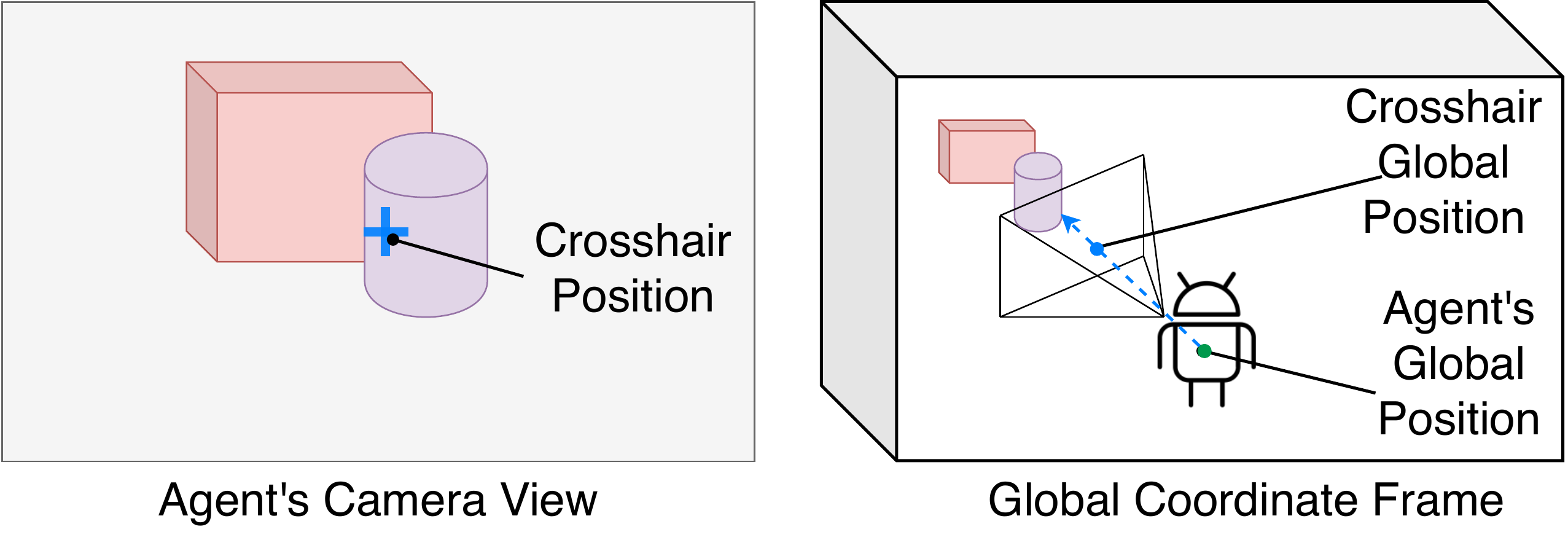}
    \caption{The grab/release action in Habitat uses the `magic pointer' abstraction to pick up objects within range.
    Any object under a fixed crosshair in the agent’s viewport can be picked by the agent if it is within a certain distance threshold.}
    \label{action_illustration}
\end{figure}

\paragraph{Action Space.}  The action space for the rearrangement task consists of navigation and interactive actions. Navigation actions includes \texttt{move\_forward} 0.25m, \texttt{turn\_left} $10\degree$, \texttt{turn\_right} $10\degree$ and \texttt{stop}. Interactive action \texttt{grab\_release} uses the magic pointer abstraction discussed earlier to pick nearby objects that are visible in the agent's field of view. Specifically, any object under a fixed crosshair in the agent's viewport can be picked by the agent if it is within a certain distance threshold. As illustrated in Figure~\ref{action_illustration}, this action works by tracing a 3D ray from the camera to the crosshair position in the near-plane of the viewing frustum and extending it until it hits a object or the distance threshold is reached. The object that intersects the ray is picked. For this scenario, the crosshair position is located at 128x176 in a 256x256 viewport and the distance threshold is 1.0m. The \texttt{grab\_release} action put the object in an invisible backpack. The agent can only carry one object at a time and calling the \texttt{grab\_release} action will release the object and put it back at the same relative location \wrt to the agent as it was picked.

\paragraph{Metric.}
Following Section~\ref{sec:evaluation}, we use
task completion as the primary metric. Specifically, an object is
considered to have been rearranged successfully if it is placed within $1m$ of its
desired goal location (as measured by the distance between the center of mass
of the object in the desired and final pose). Task completion is the percentage of goal objects rearranged successfully.

We also report episode-level success --
an episode is considered successful ($S=1$) if all objects specified in that episode
are placed correctly. This episodic success metric can be useful in measuring
the combinatorial planning aspects of the problem; for instance, if certain objects simply cannot
be successfully placed in their goal locations without first moving another object.
However, it is also noisier and thus is not considered the primary metric.

To measure how efficiently the agent performed the task, we measure Episodic \textit{Success Weighted by Path Length (SPL)} using the length of the shortest-path trajectory $l$ and the length of an agent's path $l_a$ for an episode. SPL is defined as $S l/\text{max}(l_a, l)$. SPL intuitively captures how closely the agent followed the shortest path and successfully completed the episode. Shortest path is computed by posing the rearrangement task as an extension of the traveling salesman problem. We use OR-Tools~\cite{ortools}, a combinatorial optimization library,
to find the solution to the generalized traveling salesman problem.

\end{document}